\documentclass{article}

\usepackage{microtype}
\usepackage{graphicx}
\usepackage{subcaption}
\usepackage{booktabs} 
\usepackage{multirow}
\usepackage{wrapfig}
\usepackage{enumitem}

\usepackage{hyperref}

\usepackage[preprint]{icml2026}

\usepackage{amsmath}
\usepackage{amssymb}
\usepackage{mathtools}
\usepackage{amsthm}

\usepackage[capitalize,noabbrev]{cleveref}

\theoremstyle{plain}

\theoremstyle{definition}

\theoremstyle{remark}

\usepackage[textsize=tiny]{todonotes}

\usepackage{xspace}

\newcommand{\pjn}{FOREST\xspace}
\newcommand{\dataset}{ARMBench\xspace}

\icmltitlerunning{Visual Foresight for Robotic Stow: A Diffusion-Based World Model from Sparse Snapshots}

\begin{document}

\twocolumn[
  \icmltitle{Visual Foresight for Robotic Stow: \\ A Diffusion-Based World Model from Sparse Snapshots}



  \icmlsetsymbol{equal}{*}

  \begin{icmlauthorlist}
    \icmlauthor{Lijun Zhang}{comp}
    \icmlauthor{Nikhil Chacko}{comp}
    \icmlauthor{Petter Nilsson}{comp}
    \icmlauthor{Ruinian Xu}{comp}
    \icmlauthor{Shantanu Thakar}{comp}
    \icmlauthor{Xibai Lou}{comp}
    \icmlauthor{Harpreet S. Sawhney}{comp}
    \icmlauthor{Zhebin Zhang}{comp}
    \icmlauthor{Mudit Agrawal}{comp}
    \icmlauthor{Bhavana Chandrashekhar}{comp}
    \icmlauthor{Aaron Parness}{comp}
  \end{icmlauthorlist}

  \icmlaffiliation{comp}{Amazon, Seattle, US}

  \icmlcorrespondingauthor{Lijun Zhang}{ljzhang@amazon.com}

  \icmlkeywords{Robotics, Diffusion Model, World Model}

  \vskip 0.3in
]



\printAffiliationsAndNotice{}  

\begin{abstract}
Automated warehouses execute millions of stow operations, where robots place objects into storage bins. 
For these systems it is valuable to anticipate how a bin will look from the current observations and the planned stow behavior before real execution. 
We propose \pjn, a stow-intent-conditioned world model that represents bin states as item-aligned instance masks and uses a latent diffusion transformer to predict the post-stow configuration from the observed context.
Our evaluation shows that \pjn substantially improves the geometric agreement between predicted and true post-stow layouts compared with heuristic baselines. 
We further evaluate the predicted post-stow layouts in two downstream tasks, in which replacing the real post-stow masks with \pjn predictions causes only modest performance loss in load-quality assessment and multi-stow reasoning, indicating that our model can provide useful foresight signals for warehouse planning.
\end{abstract}

\section{Introduction} \label{sec:intro}
Stow is the operation of placing objects into storage containers such as bins or shelves~\cite{cosgun2011push, basu2012learning, chen2023predicting}. 
Humans perform similar actions when organizing books on a bookshelf or loading a cabinet. 
In large fulfillment centers, the same operation must be carried out millions of times per day, making manual execution costly and limiting overall throughput. 
As a result, advanced warehouses deploy robotic systems that automate the stow process. 
Stow has become an important task in robotic manipulation aimed at improving storage efficiency while maintaining inventory accuracy~\cite{hudson2025stow, park2025pick}.

In this work we study a complementary capability for robotic stow that we call \emph{visual foresight for stow}. 
Visual foresight broadly denotes predicting how future visual observations will look given a current state and an intended physical action, and is closely related to world models in robotics and control~\cite{finn2017visualforesight, ebert2018visual, bar2025navigation}. 
Such predictive models do not replace but complement planners by providing useful supervision about dynamics that serve as signals when choosing from candidate actions. 
In this paper, we consider the visual foresight problem in a bin-based stow system. 
Given a pre-stow bin observation, the item to be stowed, and a high-level description of how the robot intends to place the item, we wish to forecast the bin layout after the specified stow, before physical execution. 
Accurate forecasts can support downstream decision making, for example by estimating how a proposed stow will affect storage density or by performing long-horizon planning.

\begin{figure}[htb]
    \vspace{-8pt}
    \centering
    \includegraphics[width=.45\textwidth]{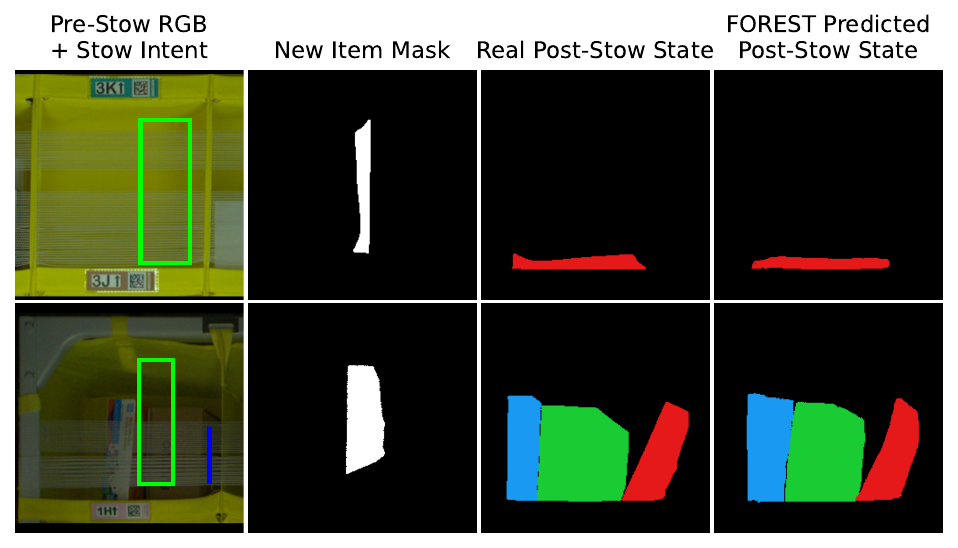}
    \vspace{-.2cm}
    \caption{Examples of visual foresight for robotic stow.}
    \label{fig:start_vis}
    \vspace{-8pt}
\end{figure}

Studying visual foresight for stow in a real production poses several challenges. 
First, supervision is inherently sparse because continuous video capture at warehouse scale is impractical, and intermediate motion is frequently occluded by the robot and the incoming item. 
Consequently, production datasets commonly provide only pre- and post-stow RGB snapshots, as in \dataset~\cite{hudson2025stow}. 
Second, real deployments exhibit substantial diversity, including thousands of distinct items that vary in size, shape, and rigidity, as well as storage bins with different geometries, which jointly affect perception and contact outcomes. 
Third, stow is interaction intensive because the incoming item can push, slide, or topple existing items and induce non-local rearrangements that are difficult to infer from sparse observations (see Figures~\ref{fig:start_vis}). 
These factors motivate the world-modeling framework proposed in this work for learning object dynamics in real stow scenarios.

Given these challenges, we approach visual foresight for stow as learning a stow-intent-conditioned world model and propose \pjn (FOREST, \underline{Fore}sight for \underline{ST}ow).
Starting from the sparse pre- and post-stow snapshots, we first derive a structured prediction target by extracting instance masks and aligning items before and after stow via item matching, so that each physical item occupies a dedicated slot in both the pre- and post-stow bin states.
\pjn then learns to map the pre-stow bin state, a new-item observation, and a high-level stow intent to a predicted post-stow bin state in this slot-based representation.
Specifically, a transformer-based latent diffusion model, conditioned on the stow context, is designed to approximate the item-level rearrangements induced by real stow operations.

Our main contributions are:
\begin{itemize}[noitemsep,nolistsep,topsep=0pt]
    \item \emph{\pjn} --- a framework for visual foresight of robotic stow from sparse pre- and post-stow snapshots in the real production. 
    To the best of our knowledge, \pjn is the first learned world model that operate in real stow scenario with only snapshot supervision.

    \item \emph{Direct evaluation} --- we quantify post-stow prediction performance using instance-level IoU between the predicted and ground-truth post-stow masks. 
    \pjn consistently improves IoU for the newly inserted item over heuristic baselines by $0.3$-$0.5$, while maintaining high IoU for pre-existing items.

    \item \emph{Downstream evaluation} --- we further evaluate whether predicted post-stow masks are effective inputs for two downstream tasks. 
    For bin free-space change prediction, which measures the change in available horizontal stow space and serves as a load-related proxy, replacing ground-truth post-stow masks with \pjn-predicted masks increases prediction MAE by only $0.0016$-$0.0025$. 
    For multi-stow reasoning, we roll out \pjn over consecutive stows in the same bin and show that it supports long-horizon prediction.
\end{itemize}

\section{Problem Statement and Preliminaries} \label{sec:preliminary}
We first introduce the targeted stow data, formalize the visual foresight problem for robotic stow, and then review the diffusion model background necessary for our world model.
Additional related work is provided in Appendix~\ref{sec:related}.

\begin{figure}[htb]
    \vspace{-9pt}
    \centering
    \includegraphics[width=.4\textwidth]{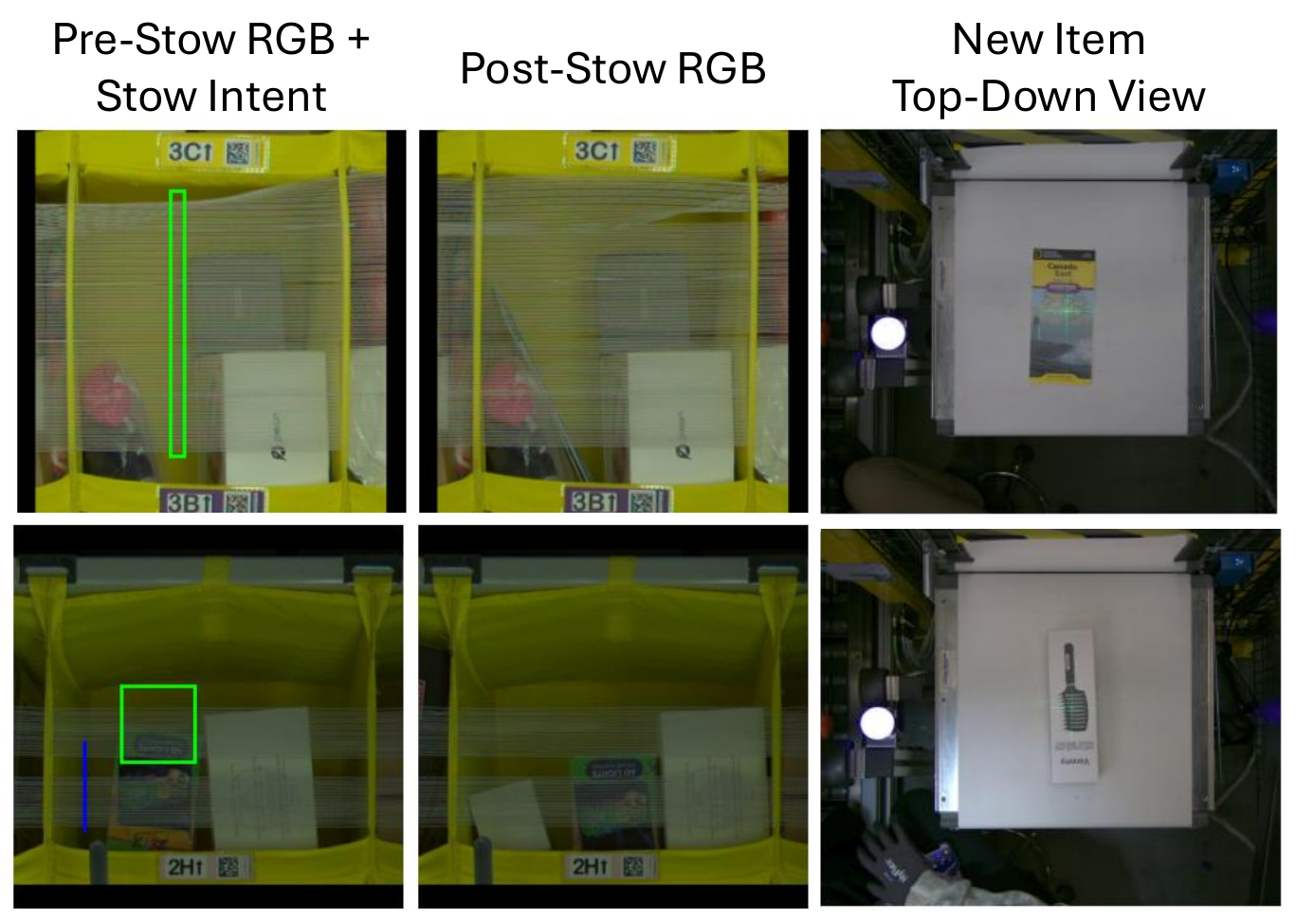}
    \vspace{-.1cm}
    \caption{Examples in \dataset with pre-stow RGB, post-stow RGB, and induct-view new item image. The stow intents are depicted by bounding box overlayed on the pre-stow RGB image, in which green one refers to the planned position for new item placement while the blue one is the location of sweeping operation.
    }
    \label{fig:dataset}
    \vspace{-8pt}
\end{figure}

\textbf{Stow Data in Production.}
We target \dataset~\cite{hudson2025stow}, a large-scale dataset of real stows collected from robotic warehousing systems in production (see Figure~\ref{fig:dataset}).
Specifically, items are stored in rectangular fabric bins observed by a frontal camera, and a \emph{stow event} corresponds to placing one item into a target bin.
For each stow event, \dataset provides:
(i) a pre-stow RGB image of the bin,
(ii) a post-stow RGB image of the same bin,
(iii) an top-down RGB image of the incoming item captured at the induction station,
(iv) the incoming item’s physical properties, and
(v) the stow intent, that is, high-level metadata indicating how the robot plans to execute the stow.

The system follows a bookshelf-like stowing paradigm in which item side faces are generally oriented toward the camera for subsequent picking~\cite{hudson2025stow, park2025pick}. 
Accordingly, the stow intent specifies a planned placement pose and, when necessary, a sweeping operation that first creates free space by pushing existing items before inserting the new item. 
This induces an unobserved intermediate state between the pre- and post-stow snapshots, making post-stow state prediction particularly difficult. 
Also despite this coarse intent taxonomy, the task remains challenging due to the massive variety of incoming items and the use of deformable fabric bins, whose effective geometry and available space can change under contact.


\textbf{Visual Foresight for Stow.}
We view visual foresight for stow as learning a transition model over bin states. Given the bin state before a stow, an observation of the new item, and the planned stow intent, the model should anticipate how the bin state will change after executing that stow.
Unlike many prior visual foresight settings that rely on dense video sequences~\cite{finn2017visualforesight,ebert2018visual}, each stow event in \dataset is observed only through a pair of temporally sparse snapshots, a pre-stow bin state and a post-stow bin state. 
Our goal is to learn from these sparse observations a single-step world model~\cite{ha2018world} that predicts a structured future bin state.

Formally, let $\mathcal{X}$ denote the space of bin states, $\mathcal{O}$ the space of new-item observations, and $\mathcal{U}$ the space of stow intents.
Each stow event is represented as a tuple $(x_{\text{pre}}, o_{\text{new}}, u, x_{\text{post}})$, where $x_{\text{pre}} \in \mathcal{X}$ is the pre-stow bin state, $o_{\text{new}} \in \mathcal{O}$ is the observation of the item to be stowed, $u \in \mathcal{U}$ is the associated stow intent, and $x_{\text{post}} \in \mathcal{X}$ is the post-stow bin state.
Given a dataset $\mathcal{D} = \{(x_{\text{pre}}^{(i)}, o_{\text{new}}^{(i)}, u^{(i)}, x_{\text{post}}^{(i)})\}_{i=1}^N$, we aim to learn a world model $F_\theta : \mathcal{X} \times \mathcal{O} \times \mathcal{U} \rightarrow \mathcal{X}$ that maps the current state, new item, and intent to a prediction of the future state:
\begin{equation}
  \hat{x}_{\text{post}}
  = F_\theta\big(x_{\text{pre}},\, o_{\text{new}},\, u\big).
\end{equation}
In \S\ref{sec:method}, we instantiate $x_{\text{pre}}$ and $x_{\text{post}}$ with an instance-mask-based representation of bin states, and $F_\theta$ with a diffusion-based world model.

\textbf{Diffusion-Based World Models.}
Diffusion models~\cite{sohl2015deep} are generative models that synthesize data by learning to invert a gradual noising process. We summarize the basics in Appendix~\ref{app:diffusion}.
For high-resolution vision tasks, it is effective to apply diffusion in a compact latent space rather than directly in pixel space.
For instance, Stable Diffusion~\cite{rombach2022high} operates on the latent space of a pre-trained variational autoencoder (VAE).
Beyond image synthesis, recent work show that diffusion models can serve as world models, learning predictive dynamics that map current states and actions to future states in domains such as games, autonomous driving, and robotic manipulation~\cite{alonso2024atari, gao2024vista, huang2025ladi}.
Motivated by these findings, we instantiate visual foresight for stow as a latent diffusion world model conditioned on the stow context, as detailed in \S\ref{sec:diffusion_arch}.

\section{\pjn: A Diffusion-Based World Model for Visual Foresight in Stow} \label{sec:method}

\subsection{Overview of \pjn} \label{sec:method_overview}
\begin{figure*}[htb]
    \centering
    \includegraphics[width=.93\textwidth]{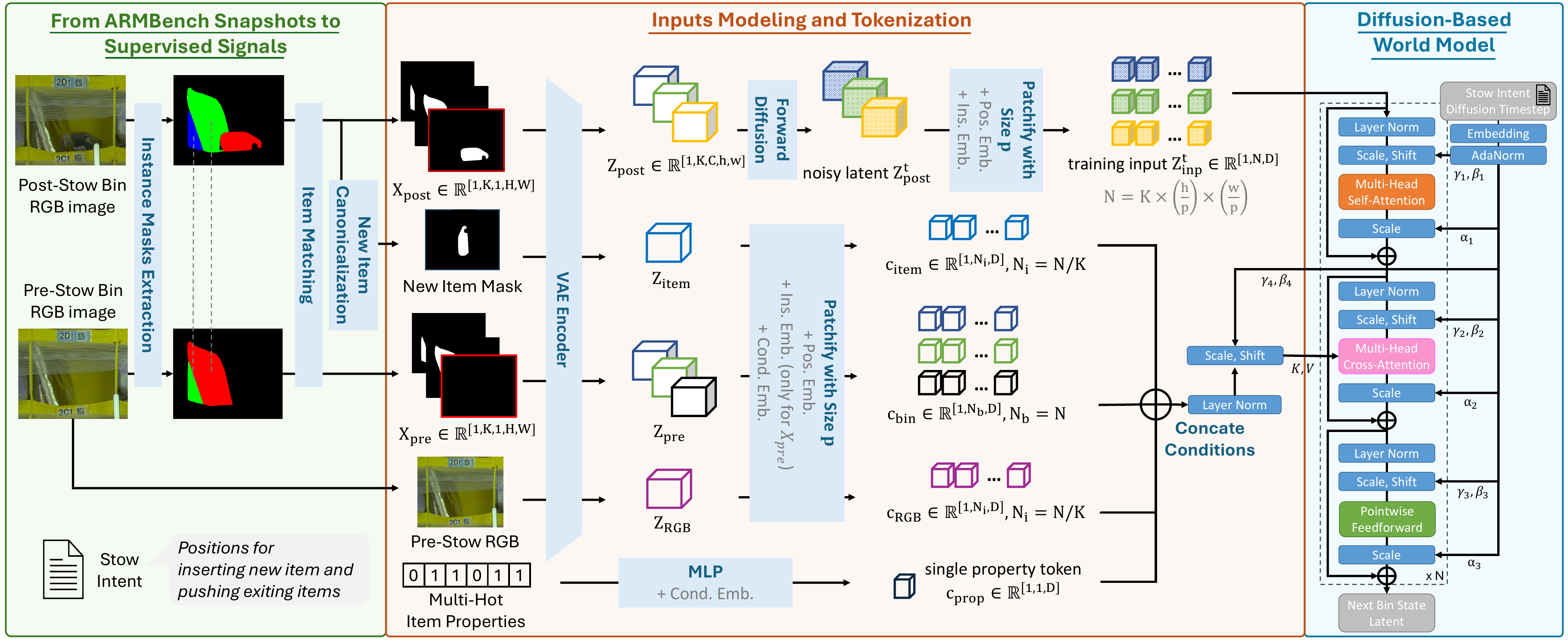}
    \vspace{-.2cm}
    \caption{
        Overview of \pjn, our diffusion-based world model for visual foresight of stow with three stages.
        Stage~1 constructs slot-aligned pre- and post-stow bin states from \dataset snapshots via instance masks extraction and item matching.
        Stage~2 encodes bin states and stow context into latent vectors and then noisy tokens and condition tokens.
        Stage~3 uses a transformer-based latent diffusion model with cross-attention and adaptive layer normalization to incorporate condition tokens.
        }
    \label{fig:overview}
    \vspace{-5pt}
\end{figure*}

Figure~\ref{fig:overview} summarizes the overall pipeline of \pjn with three stages.
We first explain the high-level idea, and then detail each stage.

\textbf{Stage 1: From \dataset Snapshots to Supervised Signals.}
We first extract instance masks for all items in the pre- and post-stow RGB snapshots of each stow event. 
An item-matching procedure then establishes a one-to-one correspondence between pre- and post-stow instances. 
Stacking the binary masks according to this mapping yields slot-based pre- and post-stow bin states that describe how items occupy the bin before and after the operation. The unmapped extra mask in the post-stow instances is the mask of the newly stowed item and is canonicalized before being fed into the model.
These item-aligned bin states and the identified new item mask serve as supervision for training the world model.
\S\ref{sec:item_matching} details mask extraction and item matching steps.

\textbf{Stage 2: Input Modeling and Tokenization.}
Next, we encode both the stow context and the future bin layout into latent features and tokens. 
An off-the-shelf VAE~\cite{vae} transforms inputs, including the slot-based pre- and post-stow bin states, the canonical item mask, and the pre-stow snapshot, from either binary masks or RGB image to latent feature maps.
For the post-stow latent, we sample a diffusion timestep, add Gaussian noise according to the diffusion forward process, and patchify the noisy latent into a sequence of input tokens during training. 
The other latents as well as the conditioned item property embedding are structured into condition tokens, augmenting with learned instance and condition-type embeddings such that the model can distinguish different items and conditioning sources.
\S\ref{sec:input_modeling} presents this modeling and tokenization step.

\textbf{Stage 3: Diffusion-based World Model.}
Finally, we instantiate \pjn as a transformer-based latent diffusion model. 
At each diffusion timestep, the model takes the noisy post-stow tokens as queries and the condition tokens from the stow context as keys and values in cross-attention layers, so that the denoising updates depend explicitly on the pre-stow configuration and the new item information. 
A joint embedding of the stow intent and diffusion timestep modulates every transformer block through Adaptive Layer Normalization (AdaLN)~\cite{xu2019adaln}.
\S\ref{sec:diffusion_arch} describes the architecture in detail.

\subsection{From \dataset Snapshots to Supervised Signals} \label{sec:item_matching}

\pjn is trained from raw \dataset streams by converting each stow event into (i) a pair of slot-aligned pre- and post-stow bin states and (ii) a canonical new-item mask used as the item observation. 

\textbf{Instance Masks Extraction.}
Since our goal is to predict how a stow rearranges the bin at the object level, we represent bin states as sets of binary instance masks, one per visible item. 
This object-centric representation makes item contacts explicit while avoiding the need to model fine-grained textures that are less relevant to stow outcomes. 
As \dataset does not provide instance masks, we derive from the RGB images using a MaskDINO-based instance segmentation model~\cite{li2022mask}, and use the resulting masks as the starting point for bin-state construction.

\textbf{Item Matching.}
Based on the extracted instance masks, we construct slot-based, item-aligned bin states via item matching.
For each stow event, since we focus on single-transition cases where exactly one new item is stowed into the bin, the numbers of instances in the pre-and post-stow snapshots satisfy $N_{\text{post}} = N_{\text{pre}} + 1$.
In each snapshot, we first stack the binary instance masks as channels and refer to each channel as a \emph{slot}, so that each slot corresponds to a single physical item.
An item-matching procedure then establishes a one-to-one correspondence between pre- and post-stow instance slots such that every pre-existing item, visible in both snapshots, is assigned the same slot index before and after the stow, and the remaining post-stow instance is identified as the newly stowed item.

We formulate item matching as an assignment problem with priors derived from the stow behavior.
Let $\{m_i^{\text{pre}}\}_{i=1}^{N_{\text{pre}}}$ and $\{m_j^{\text{post}}\}_{j=1}^{N_{\text{post}}}$ denote the pre- and post-stow instance masks.
For each mask, we crop the RGB patch and extract a visual embedding with pretrained DINOv2~\cite{oquab2023dinov2}, obtaining $\{e_i^{\text{pre}}\}$ and $\{e_j^{\text{post}}\}$, and build a base cost matrix from their cosine similarity $D_{ij}$.
We then add two priors derived from the stow intent, a placement prior that favors assignments placing the new item near the planned insertion position and, when present, the sweeping position, and an order-preservation prior that discourages changes in the left-to-right ordering of pre-existing items.
The final cost is
\begin{equation}
  C_{ij}
  = (1 - D_{ij})
  + \lambda_{\text{pos}} \, \Phi_{\text{pos}}(i,j)
  + \lambda_{\text{ord}} \, \Phi_{\text{ord}}(i,j),
\end{equation}
where $\Phi_{\text{pos}}$ and $\Phi_{\text{ord}}$ encode placement and order penalties (see details in Appendix~\ref{app:method_details}).
We solve this assignment with Hungarian algorithm~\cite{kuhn1955hungarian} to obtain a one-to-one matching consistent with visual similarity and stow priors.

Given the matching result, we reserve the first slot for the newly stowed item and assign all matched pre-existing items to the remaining slots.
In particular, the first slot is set to all zeros for the pre-stow bin state, reflecting that the new item is not yet present.
These slot-aligned bin states serve as the supervision signal for training the world model.

\textbf{New Item Canonicalization.}
\dataset provides a separate induct-view image for the new item, however, the post-stow bin layout is more directly related to the in-bin contact-surface view recovered from item matching.
We therefore rely on the post-stow instance mask of the new item as the basis for the new-item representation.
Particularly, the mask is canonicalized to avoid leaking post-stow pose and location information into the model input. 
We define a canonical pose and location by translating the new-item mask to the center of a fixed-size canvas and rotating it so that its principal axis is upright.
The resulting canonical binary mask serves as the new-item observation to \pjn.

\textbf{Clarifications.}
Instance-mask extraction and item matching can introduce noisy supervision. 
We mitigate obvious failures by filtering out stows whose post-stow instance count does not equal the pre-stow count plus one, and by discarding stows whose matches disagree with an automatic consistency check from large vision–language model based on the pre- and post-stow RGB images.
In addition, the canonical new-item mask extracted from the ground-truth post-stow bin state is used in both training and evaluation, so that we can isolate the quality of the predicted post-stow layouts under the assumption that an in-bin contact-surface view of the new item is available (see details in Appendix~\ref{app:method_details}).

\subsection{Input Modeling and Tokenization}
\label{sec:input_modeling}
We now describe how we represent the bin state, the new item, and the stow intent as latent tokens or embeddings that serve as inputs and conditions for the diffusion model.

\textbf{Bin State Tokens.}
We construct pre- and post-stow bin state tokens from the slot-based, item-aligned binary masks.
We first apply the off-the-shelf VAE encoder on each item slot, independently transforming the multi-channel binary masks into latent feature maps.
Then each latent feature map is patchified into a sequence of tokens with 2D positional embeddings to preserve spatial structure within each slot.
We further introduce two types of additional embeddings that are added to the token features.

First, each item slot is associated with a learnable \emph{instance embedding}, which is added to all tokens originating from that slot so that the model can distinguish different items and track across layers.
When training, we randomly shuffle the slots of the pre-existing items, and rely on the instance embeddings to emphasize the one-to-one correspondence.
Second, the tokens from pre-stow bin state is associated with a \emph{condition-type embedding} that indicates whether a token represents a bin state or other conditioning sources.

We also include tokens generated from the pre-stow RGB image, encoded by the same VAE, patchified into tokens, and augmented with positional and condition-type embeddings, as additional context.
This extra stream provides richer visual information about the storage bin (see ablation study in Appendix~\ref{sec:exp_ablations}).

\textbf{New Item Tokens.}
We construct new item tokens from the canonical new-item mask.
The processing mirrors that of the bin state tokens with positional and condition-type embeddings.
\dataset also provides discrete item properties, which we encode as a multi-hot vector over six attributes, rigid, round, square, conveyable, foldable, and fragile.
This multi-hot vector is embedded into a single property token using a two-layer MLP and is augmented with condition-type embedding.
The contribution of the property token is demonstrated in Appendix~\ref{sec:exp_ablations}.

\textbf{Stow Intent Embeddings.}
The stow intent describes how the robot plans to execute the stow.
As discussed in \S\ref{sec:preliminary}, the system follows a vertical, bookshelf-like stowing strategy, and the intent includes the planned position for placing the new item and, when necessary, a sweeping operation used to create additional free space. Both operations are represented as bounding boxes in the bin coordinate frame. 
Formally, we denote the placement intent by $u_{\text{place}} \in \mathbb{R}^4$ and the sweeping intent, when present, by $u_{\text{push}} \in \mathbb{R}^4$, each encoding $(x_1, y_1, x_2, y_2)$.
To condition on the stow intent $u = \{u_{\text{place}}, u_{\text{push}}\}$, we first map each scalar coordinate to a $d/4$-dimensional sine–cosine feature, concatenate the four features into a $d$-dimensional vector, and then apply a two-layer MLP to obtain intent embeddings $\Psi_{u_{\text{place}}}, \Psi_{u_{\text{push}}} \in \mathbb{R}^d$.
We follow a similar procedure to map the diffusion timestep $t \in \mathbb{R}$ to a timestep embedding $\Psi_t \in \mathbb{R}^d$.
Finally, we sum all embeddings into a single conditioning vector
\begin{equation}\label{eq:adaln-condition}
    \xi = \Psi_t + \Psi_{u_{\text{place}}} + \Psi_{u_{\text{push}}},
\end{equation}
which is used to modulate every transformer block via AdaLN in the diffusion architecture described in \S\ref{sec:diffusion_arch}.

\subsection{Diffusion-Based World Model Architecture} \label{sec:diffusion_arch}
We now describe how the tokens and embeddings are used in a diffusion-based world model that predicts the post-stow bin state in latent space.
Our architecture follows the latent diffusion with a transformer backbone with the main design choices lying in how we inject stow context and how we handle variable-length bin layouts.
Training and sampling follow the conventional noise-prediction objective and denoising procedure used in prior latent diffusion models~\cite{rombach2022high} as described in Appendix~\ref{app:diffusion}.

\textbf{Conditioning Mechanism.}
At each diffusion timestep $t$, the model operates on two token sequences: (i) the noisy tokens from the diffused post-stow latent, and (ii) the condition tokens, i.e., the pre-stow tokens and the new item tokens.
Within each transformer block, we use cross-attention to incorporate these condition tokens into the denoising update. Specifically, the noisy post-stow tokens are used as queries, and all condition tokens are used as keys and values, so that the updated latent representation of each post-stow patch is computed explicitly as a function of the pre-stow bin configuration and the new item.

To inject intent information, the conditioning vector $\xi$ in Eq.~\ref{eq:adaln-condition} modulates every transformer block through AdaLN~\cite{xu2019adaln}, providing a global signal without introducing additional intent tokens into the attention layers.

\textbf{Varied-Length Attention.}
Because different stow events contain different numbers of pre-existing items in pre-stow bins, the number of tokens per stow varies with the number of occupied slots.
Naively padding all sequences to the same length would introduce many uninformative tokens and waste computation. Instead, we adopt a packed-sequence attention scheme.
For a mini-batch of stows, we concatenate all tokens along the slot dimension and construct a block-diagonal attention mask~\cite{Dao2022FlashAttentionFA} that restricts attention to tokens belonging to the same stow.
This allows all tokens to be processed in a single batched attention operation, while preserving independence across stows and avoiding padding overhead.


\section{Empirical Evaluation} \label{sec:experiments}

\subsection{Experimental Setup} \label{sec:exp_setup}
\textbf{Datasets.}
We conduct all experiments on \dataset.
We retain only successful stow events, so that post-stow bin layouts provide reliable supervision, and group them into two intent modes, \textbf{direct insert}, where only insertion is used, and \textbf{sweep insert}, where an additional sweeping operation creates free space.
For each mode, we randomly split stows into train and test sets with an 8:2 ratio. The resulting counts are reported in Appendix~\ref{app:exp_settings}.

\textbf{Baselines and Comparisons.}
To the best of our knowledge, \pjn is the first method to address visual foresight for stow from temporally sparse pre- and post-stow snapshots rather than dense videos.
Because there is no existing baseline in this setting, we compare against heuristic baselines and design both direct and downstream evaluations to obtain a comprehensive assessment.
The \textbf{copy-paste} baseline places the new item mask into the pre-stow bin state at the intended insertion region without modeling interactions with existing items.
The \textbf{copy-paste with gravity} baseline further applies a vertical settling step that moves the pasted item downward until it hits the bin ground, approximating a simple drop of the new item into the bin.
The direct evaluation then compares the synthetic post-stow bin state with the ground-truth in terms of binary instance masks.

We also ask whether predicted post-stow bin states are useful for downstream tasks. 
Our first task is \textbf{Delta Linear Opportunity (DLO) prediction}, which measures how much usable linear space in the bin changes after a stow.
In \dataset~\cite{hudson2025stow}, Linear Opportunity (LO) is a scalar that represents the remaining available space in the horizontal direction of the bin in meters, and DLO is the difference between post- and pre-stow LO. In production, observing DLO only after execution is mainly diagnostic, while forecasting DLO for candidate stows can help choose among alternative policies.

Predicted post-stow layouts are a natural input for such forecasts, since DLO depends on how the bin changes between the pre- and post-stow states.
However, if we trained a DLO predictor directly on \pjn outputs, the error would mix the quality of the synthetic masks with the capacity of the DLO model itself.
Instead, we train a vision-based DLO predictor on ground-truth post-stow masks (see architecture in Appendix~\ref{app:exp_settings}).
The predictor takes as inputs the pre-stow RGB image, pre-stow instance masks, and a post-stow instance mask.
During training, this post-stow mask is always the ground-truth one, so the network learns the best achievable mapping from true pre and post states to DLO, independent of any visual foresight method.
At test time, we freeze this predictor and keep the pre-stow inputs fixed, but replace the post-stow mask with either the \pjn-predicted mask or the mask from the copy-paste baselines.
Differences in DLO prediction accuracy can then be attributed purely to the quality of the synthetic post-stow masks, and thus quantify how well visual foresight can stand in for ground-truth supervision in this downstream task.

The second downstream task is \textbf{multi-stow reasoning}, which evaluates the ability of \pjn to support long-horizon forecasting over multiple stows in the same bin.
We link individual stows in \dataset into chains based on bin identifiers and timestamps, forming 1,581 multi-stow sequences with chain length up to four.
Then the visual foresight model is applied sequentially along these chains, where the predicted post-stow bin state from one step becomes the pre-stow bin state for the next.
Since \pjn also requires a pre-stow RGB image as input, we synthesize the post-stow RGB image by rendering the new item into the bin using its induct view texture and the predicted mask. The synthesized post-stow RGB image then serves as the pre-stow RGB input for the next stow in the chain. Details of this synthesis procedure are in Appendix~\ref{app:exp_settings}.

We consider two evaluation modes for multi-stow reasoning.
In the \emph{rollout} mode, the predicted post-stow bin state at each step are fed into the next step, so that errors can accumulate over time.
In the \emph{teacher forcing} mode, we instead provide ground-truth pre-stow bin states at every step, without feeding previous predictions forward.

\textbf{Metrics.}
For direct evaluation of post-stow bin states, we report instance-level intersection-over-union (IoU) for both the newly inserted item (N-IoU) and pre-existing items (O-IoU).
And we further perform two fine-grained analyses.
We first measure the size of the new item and partition stow events into small, medium, large buckets based on quantiles of the item size distribution and report N-IoU respectively.
We also explore performance regarding stow events difficulty. Specifically, we use the N-IoU achieved by the copy-paste baseline as a proxy for how surprising or obvious a stow is.
Details of the stow partitions are provided in Appendix~\ref{app:exp_settings}.

For DLO prediction, we report the mean absolute error (MAE) between the predicted and ground-truth DLO.
Using ground-truth post-stow masks gives an upper bound on the achievable performance of the trained DLO predictor, while the increase in MAE when using synthetic post-stow masks measures the loss due to imperfect post-stow predictions.

In addition, we assess prediction-level agreement via a linear regression between DLO obtained with ground-truth masks and with synthetic masks.
For each setting, we fit a line $y = a + b x$, where $x$ is the DLO predicted using ground-truth post-stow masks and $y$ is the DLO predicted using synthetic masks from either \pjn or the baselines.
We report the slope $b$, the intercept $a$, and the coefficient of determination $R^2$, defined as
\begin{equation}
    R^2
    = 1 - \frac{\sum_i (y_i - \hat{y}_i)^2}
             {\sum_i (y_i - \bar{y})^2},
\end{equation}
where $y_i$ is the synthetic-mask DLO prediction, $\hat{y}_i$ is its regression fit, and $\bar{y}$ is the mean of $\{y_i\}$.
The ideal case, where synthetic-mask predictions exactly match ground-truth–based predictions, would have $a = 0$, $b = 1$, and $R^2 = 1$.

For multi-stow reasoning, we report N-IoU at each step under both teacher-forcing and rollout modes, and use their gap as a measure of accumulated error over longer horizons.

\textbf{\pjn Settings.}
\pjn has 50M parameters and we evaluate two variants. \textbf{\pjn-DI} and \textbf{\pjn-SI} are trained separately on direct-insert and bin-sweep stows, while \textbf{\pjn-J} is trained jointly on both modes. In the joint setting, the architecture adapts to different stow intents through the intent-conditioned AdaLN layers described in \S\ref{sec:diffusion_arch}.
The models are evaluated with 50 denoising steps.

\subsection{Direct Evaluation} \label{sec:exp_direct}

\textbf{Quantitative Results.}
Table~\ref{tab:post_iou} summarizes instance-level IoU between predicted and ground-truth post-stow bin states. 
Overall, \pjn markedly improves N-IoU over copy-paste baselines in both stow modes. 
On direct insert, N-IoU rises from $0.28$–$0.36$ for copy-paste to about $0.70$ for \pjn-DI and \pjn-J, while on sweep insert it increases from $0.12$–$0.22$ to $0.62$–$0.64$.

We highlight two observations.
First, small items are intrinsically harder to predict, since even a minor spatial deviation leads to a large drop in IoU. This is reflected in the baselines, whose N-IoU for small items is only $0.11$ for direct insert and $0.05$ for sweep insert. 
Despite this difficulty, \pjn attains N-IoU above $0.5$ for small items in both scenarios, and further improves to around $0.73$–$0.80$ for medium and large items.
Second, the difficulty-based breakdown clarifies \pjn's advantage in solving hard cases. 
For surprising stows, where copy-paste often fails with N-IoU near zero, \pjn raises N-IoU to around $0.56$–$0.61$, while maintaining competitive performance on the obvious stows. 

Besides, for pre-existing items, the heuristic baselines are able to achieve high O-IoU, which is consistent with the stow process. In direct insert, pre-existing items are rarely disturbed, and even when a sweeping operation is present, their relative ordering and approximate positions remain stable.
In such cases, simple copy-paste can already reproduce much of the pre-existing layout.
Nevertheless, \pjn matches or slightly improves O-IoU while delivering much larger gains on N-IoU.

\begin{table}[htb]
\centering
\caption{IoU between the predicted and the ground-truth post-stow bin states in instance-mask space. We report N-IoU for the newly inserted item and O-IoU for pre-existing items, along with fine-grained N-IoU analysis by new-item size and stow difficulty.}
\label{tab:post_iou}
\scriptsize
\tabcolsep=0.1cm
\vspace{-4pt}
\begin{tabular}{llcccc}
\toprule
Mode & Metric & Copy-Paste & \begin{tabular}[c]{@{}l@{}}Copy-Paste \\ + Gravity\end{tabular} & \begin{tabular}[c]{@{}l@{}}\pjn-DI \\ / \pjn-SI\end{tabular} & \pjn-J \\
\midrule
\multirow{11}{*}{\begin{tabular}[c]{@{}l@{}}Direct\\Insert\end{tabular} }
 & \multicolumn{5}{c}{\emph{Overall}} \\ \cmidrule{2-6}
 & N-IoU        & 0.2846 & 0.3632 & 0.7017 & \textbf{0.7021} \\
 & O-IoU        & 0.8563 & 0.8563 & \textbf{0.8550} & 0.8536 \\  \cmidrule{2-6}
 & \multicolumn{5}{c}{\emph{N-IoU by item size}} \\  \cmidrule{2-6}
 & Small        & 0.1076 & 0.2284 & \textbf{0.5771} & 0.5745 \\
 & Medium       & 0.2920 & 0.3766 & 0.7263 & \textbf{0.7311} \\
 & Large        & 0.4544 & 0.4849 & \textbf{0.8021} & 0.8010 \\  \cmidrule{2-6}
 & \multicolumn{5}{c}{\emph{N-IoU by difficulty}} \\  \cmidrule{2-6}
 & Surprises    & 0.0079 & 0.1034 & \textbf{0.5623} & 0.5569 \\
 & Non-obvious  & 0.3138 & 0.3980 & 0.7354 & \textbf{0.7379} \\
 & Obvious      & 0.6996 & 0.7069 & \textbf{0.7920} & \textbf{0.7920} \\
\midrule
\multirow{11}{*}{\begin{tabular}[c]{@{}l@{}}Bin\\Sweep\end{tabular} }
 & \multicolumn{5}{c}{\emph{Overall}} \\  \cmidrule{2-6}
 & N-IoU        & 0.1214 & 0.2167 & 0.6166 & \textbf{0.6422} \\
 & O-IoU        & 0.5287 & 0.5287 & 0.6878 & \textbf{0.6906} \\  \cmidrule{2-6}
 & \multicolumn{5}{c}{\emph{N-IoU by item size}} \\  \cmidrule{2-6}
 & Small        & 0.0457 & 0.1502 & 0.4791 & \textbf{0.5108} \\
 & Medium       & 0.1212 & 0.2270 & 0.6276 & \textbf{0.6521} \\
 & Large        & 0.1974 & 0.2732 & 0.7434 & \textbf{0.7641} \\  \cmidrule{2-6}
 & \multicolumn{5}{c}{\emph{N-IoU by difficulty}} \\  \cmidrule{2-6}
 & Surprises    & 0.0065 & 0.0857 & 0.5855 & \textbf{0.6117} \\
 & Non-obvious  & 0.2181 & 0.3332 & 0.6483 & \textbf{0.6737} \\
 & Obvious      & 0.6922 & \textbf{0.7201} & 0.6394 & 0.6561 \\
\bottomrule
\end{tabular}
\vspace{-.2cm}
\end{table}

\begin{figure*}[htb]
    \centering
    \includegraphics[width=.85\textwidth]{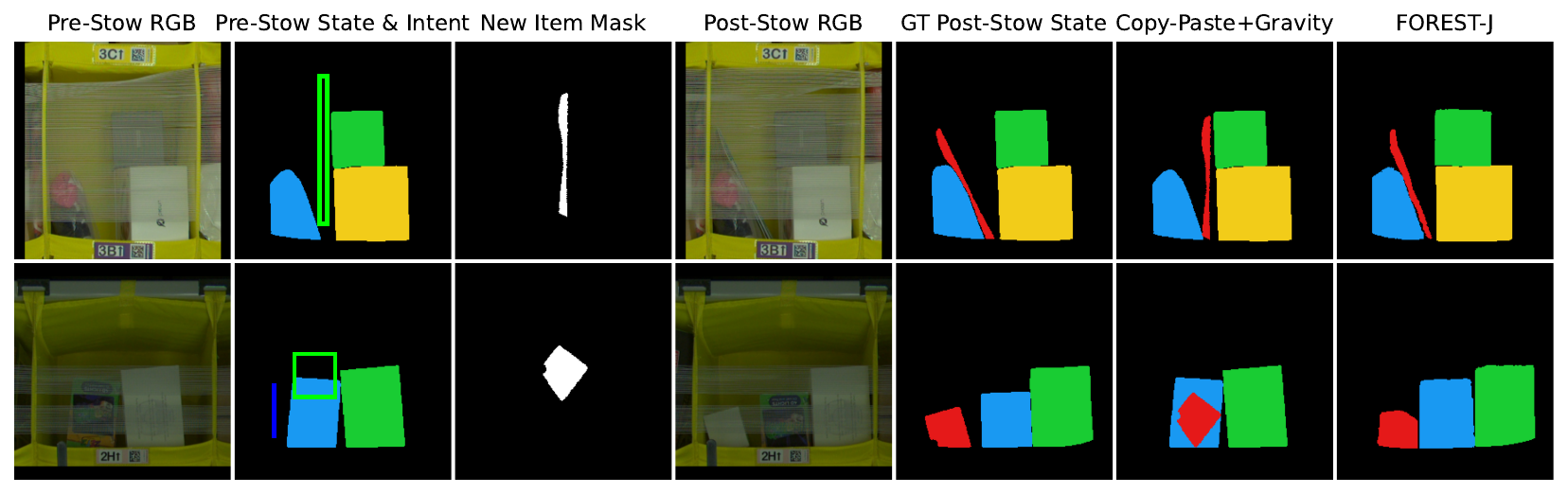}
    \vspace{-.2cm}
    \caption{Examples of post-stow bin state prediction (1st row: direct insert, 2nd row: sweep insert). }
    \label{fig:direct_vis}
    \vspace{-5pt}
\end{figure*}

\textbf{Visualizations.}
Figure~\ref{fig:direct_vis} illustrates two stow events with predicted post-stow bin states (see more visualizations in Appendix~\ref{app:vis}).
Compared to the copy-paste with gravity baseline, \pjn better captures the underlying world dynamics of the stow.
For instance, in the direct-insert case, where the new item is tall and unstable, \pjn successfully predicts the item tipping and coming to rest against existing objects, in agreement with the ground truth.
And in the bin-sweep example, where the sweeping operation shifts multiple pre-existing items to make room for the new item, \pjn anticipates both the displacement of old items and the final resting pose of the new item.
These results indicate that \pjn goes beyond learning a static placement heuristic and acquires a useful modeling of how stow actions reconfigure the bin.

\subsection{Downstream Evaluations} \label{sec:exp_downstream}

\textbf{Delta Linear Opportunity (DLO) Prediction.} 
We first study DLO prediction and compare the DLO predictor's performance when driven by different post-stow masks.

\begin{table}[htb]
\centering
\caption{MAE between predicted and ground-truth DLO when using ground-truth (GT), copy-paste with gravity (CP+G), or \pjn-predicted post-stow masks. 
} \label{tab:dlo_mae}
\scriptsize
\tabcolsep=0.15cm
\vspace{-4pt}
\begin{tabular}{c|ccc|ccc}
\toprule
Stow Intent & \multicolumn{3}{c|}{Direct Insert}                           & \multicolumn{3}{c}{Sweep Insert}                              \\
Post Mask      & GT     & CP + G & \pjn & GT    & CP + G & \pjn \\ \midrule
DLO MAE     & \textbf{0.0168} & 0.0225           & 0.0184             & \textbf{0.0255} & 0.0353             & 0.0280             \\
- GT         & -                & 0.0057            & \textbf{0.0016}    & -               & 0.0098             & \textbf{0.0025}             \\ \bottomrule
\end{tabular}
\vspace{-8pt}
\end{table}

\begin{figure}[htb]
    \centering
    \includegraphics[width=.45\linewidth]{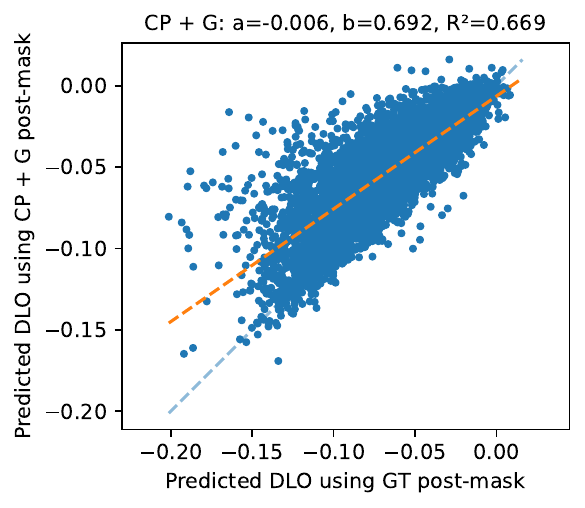}
    \includegraphics[width=.45\linewidth]{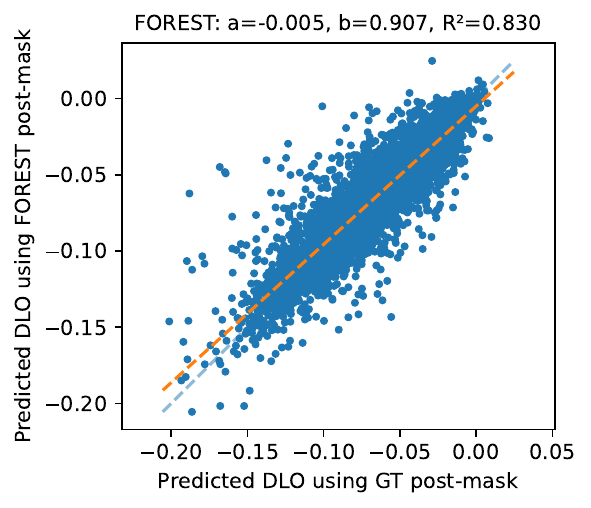} \\
    \scriptsize{(a) Direct Insert} \\
    \includegraphics[width=.45\linewidth]{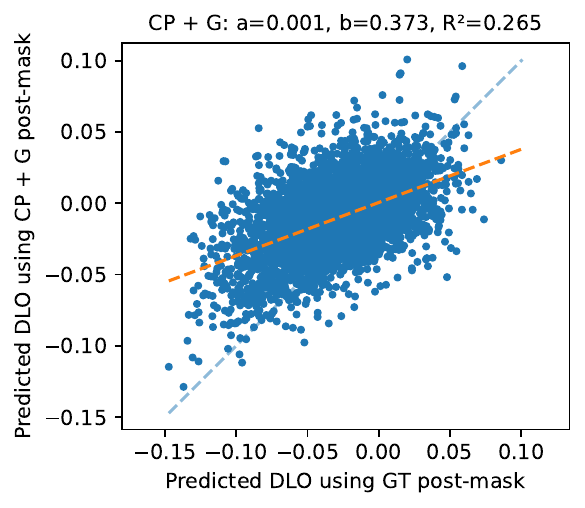}
    \includegraphics[width=.45\linewidth]{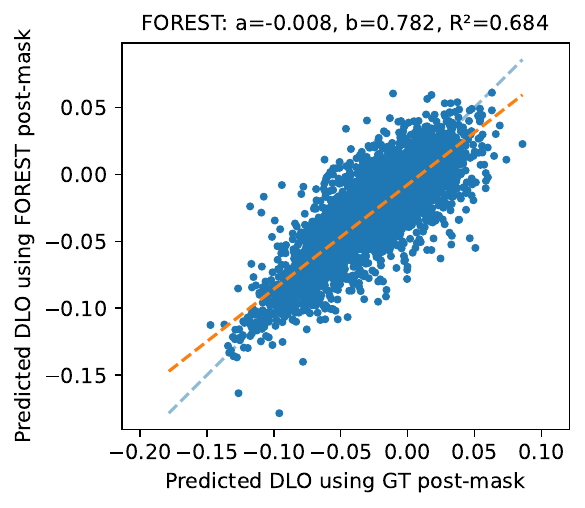} \\
    \scriptsize{(b) Sweep Insert} \\
    \vspace{-2pt}
    \caption{Linear regression between DLO predictions obtained with ground-truth (GT) post-stow masks and those obtained from copy-paste with gravity (CP+G) or from \pjn. 
    }
    \label{fig:dlo_regression}
    \vspace{-10pt}
\end{figure}

Table~\ref{tab:dlo_mae} reports the MAE between the predicted and ground-truth DLO under three choices of post-stow masks at test time, ground-truth (GT), copy-paste with gravity (CP+G), and \pjn-predicted.
Using GT post-stow masks gives the best achievable MAE for the trained DLO predictor, $0.0168$ for direct insert and $0.0255$ for sweep insert.
Replacing GT masks with CP+G masks substantially increases the error for about $0.0057$ and $0.0098$ in the two modes, reflecting the limitations of the heuristic post-stow layouts.
In contrast, using \pjn masks increases MAE by only $0.0016$ and $0.0025$, reducing the additional error by roughly $70\%$ compared to CP+G.

Figure~\ref{fig:dlo_regression} provides the linear regression between DLO predictions obtained with GT post-stow masks and those obtained with synthetic masks.
Under CP+G, the slopes are noticeably below one and $R^2$ drops to even $0.37$ in sweep insert, indicating that CP+G-based predictions deviate substantially from GT-based predictions, especially in the more challenging bin-sweep mode.
With FOREST masks, the slopes move much closer to one and $R^2$ increases to $0.83$ and $0.68$.
Together with Table~\ref{tab:dlo_mae}, these trends show that DLO predictions based on \pjn-predicted post-stow masks closely imitate those derived from ground-truth masks.

\begin{figure}[htb]
    \vspace{-.2cm}
    \centering
    \includegraphics[width=.9\linewidth]{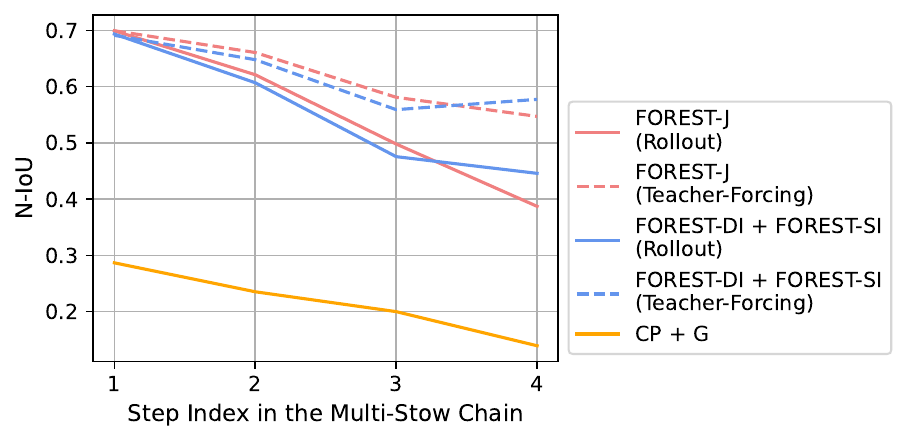}
    \vspace{-5pt}
    \caption{N-IoU of predicted post-stow states along multi-stow chains. CP+G: the copy-paste-with-gravity baseline.}
    \label{fig:multistow}
    \vspace{-.3cm}
\end{figure}

\textbf{Multi-Stow Reasoning.}
We next show that the learned world model can be rolled out over multiple stows in the same bin.
Figure~\ref{fig:multistow} reports N-IoU as a function of the step index in the multi-stow chain.
Both the joint model, \pjn-J, and the intent-specific models, \pjn-DI and \pjn-SI, start from single-step N-IoU around $0.7$ at step~1 and remain around $0.4$ at step~4, whereas the copy-paste-with-gravity (CP+G) baseline quickly degrades below $0.2$ after step~3.
These results indicate that \pjn’s predicted post-stow masks can be composed across several stows to support multi-step foresight.

We plot both teacher-forcing and rollout modes to examine how prediction errors accumulate over multiple stows.
Under teacher forcing, each step is evaluated from the ground-truth pre-stow state, so errors do not propagate across time. The mild downward trend mainly reflects that later steps occur in more cluttered, higher fill-level bins, which are intrinsically harder to predict.
Under rollout, the model instead consumes its own predicted post-stow masks and synthesized RGB as inputs, so small errors at early steps are fed back and amplified over time, widening the gap between the two curves as the horizon grows.
Per-step N-IoU numbers and additional visualizations are provided in Appendix~\ref{app:vis}.

To rule out the synthesized RGB as a confounding factor, we additionally evaluate a rollout-style setting that keeps the pre-stow RGB real at every step while still feeding predicted post-stow masks forward. The resulting curve is nearly identical to full rollout, indicating that the degradation is primarily driven by accumulated mask errors rather than RGB synthesis (see Appendix Figure~\ref{fig:multi_stow_rgb_ablation}).

\section{Conclusion}
We presented \pjn, a stow-intent-conditioned world model for robotic stow that operates on real production data. 
By representing bin states as item-aligned instance masks and applying a latent diffusion transformer conditioned on the new item and stow intent, \pjn predicts structured post-stow layouts that closely match those observed in practice.
Direct evaluation shows that \pjn improves the geometric agreement between predicted and ground-truth post-stow bin states over heuristic baselines. 
Downstream evaluations further show that substituting ground-truth post-stow masks with \pjn predictions leads to only modest degradation in load-quality prediction and multi-stow forecasting, suggesting that such world models can provide useful foresight signals for stow-related decision making.
The limitations of \pjn are discussed in Appendix~\ref{app:limitation}.

\section*{Impact Statement}

This paper presents work whose goal is to advance the field of Machine
Learning and Robotics. There are many potential societal consequences of our work, none
which we feel must be specifically highlighted here.

\bibliography{example_paper}
\bibliographystyle{icml2026}

\newpage
\appendix
\onecolumn
\section{Related Work} \label{sec:related}
\textbf{Robotic Stow Task.}
Robotic object placement has been studied in several settings, including clearing space in cluttered scenes~\cite{cosgun2011push}, using semantic cues for in-home manipulation~\cite{basu2012learning}, and predicting object motion and stability when placing items on shelves~\cite{chen2023predicting}.
The last line of work explicitly formulates a \emph{stow} task, where a robot relocates an object from a table to a cluttered shelf, closely mirroring one of the core activities in warehouse operations.
The 2015--2017 Amazon Robotics Challenges~\cite{corbato2018integrating,correll2016analysis,morrison2018cartman} further emphasized this setting by requiring teams to pick diverse items from a source tote and stow them onto shelves or into other containers.
More recently, industrial research has moved beyond challenge-style demonstrations toward deployed systems in real fulfillment centers~\cite{hudson2025stow}. The associated \dataset benchmark provides large-scale perception data from production workcells in Amazon warehouses, capturing diverse objects and cluttered container configurations.
Our work is complementary to these systems-oriented efforts. Whereas prior work focuses on building and evaluating end-to-end stow hardware and control stacks in production~\cite{hudson2025stow}, we study visual foresight for the stow task, aiming to predict post-stow bin layouts in this real production setting.

\textbf{Visual Foresight.}
Visual foresight~\cite{finn2017visualforesight} refers to learning predictive models that anticipate future visual observations conditioned on current images and candidate actions, and then using these predictions for planning. 
Early work in this area developed action-conditioned video prediction models for robot manipulation, where a convolutional network predicts future frames given past images and a sequence of actions, and downstream control chooses actions that lead predicted pixels to desired goal locations or goal images~\cite{finn2017visualforesight,ebert2018visual}.
These visual model-predictive control methods showed that self-supervised video prediction can support nonprehensile manipulation and generalize to novel objects and tasks, and were later extended to more complex settings such as tool use~\cite{xie2019tool}.
However, they typically assume dense frame observations and are designed around tabletop manipulation scenarios rather than real warehouse stow in the cluttered environment.

\textbf{World Models.}
Within reinforcement learning, visual foresight can be viewed as a special case of learning a \emph{world model}~\cite{ha2018world}, i.e., a generative model of environment dynamics that predicts future observations and rewards conditioned on current state and actions.
Typical world models employ recurrent neural networks and convolution networks to learn compact latent dynamics and then perform planning or policy learning in that latent space~\cite{ha2018world, hafner2019planet, hafner2019dream}. 
More recently, diffusion models have been adopted as expressive world models that can better capture multi-modal interactions and fine-grained visual details. 
For instance, DIAMOND introduces a diffusion-based world model for Atari, showing that modeling environment with diffusion improves visual fidelity and in turn agent performance on Atari 100k benchmarks~\cite{alonso2024atari}.
In robotic manipulation, LaDi-WM~\cite{huang2025ladi} proposes a latent diffusion world model on top of pre-trained visual foundation model features, and finds that predicting the evolution of such latent representations yields accurate future states and substantially improves policy performance in manipulation benchmarks.

Our work extends diffusion-based world models to a distinct problem, i.e., visual foresight for warehouse stow in real production systems.
In contrast to prior visual foresight and world-model approaches that typically predict dense video sequences or multi-step trajectories in laboratory environments, we study single-step transitions between pre- and post-stow snapshots observed only as temporally sparse RGB images.
The objective is to anticipate how a storage container will be reconfigured after executing a stow, conditioned on the current snapshot, the new item, and the stow intent.

\section{Diffusion Models Basis} \label{app:diffusion}
We begin by recalling the \emph{forward} diffusion process. Given a data point sampled from the real data distribution $x_0 \sim q(x)$, the forward process progressively corrupts $x_0$ into a noisy variable $x_T$ through $T$ steps of Gaussian noise addition.
In the commonly used variance-preserving formulation, this process admits a closed-form expression:
\begin{equation}\label{eq:forward-diffusion-closed}
  x_t = \sqrt{\bar{\alpha}_t}\, x_0
      + \sqrt{1 - \bar{\alpha}_t}\, \varepsilon,
\end{equation}
where $\{\bar{\alpha}_t\}_{t=1}^T$ is a deterministic noise schedule with $\bar{\alpha}_t = \prod_{i=1}^t (1 - \beta_i)$, $\beta_i \in (0, 1)$ controls the amount of noise added at step $i$, and $\varepsilon \sim \mathcal{N}(\mathbf{0}, \mathbf{I})$. As $t$ increases, the influence of $x_0$ vanishes and $x_T$ converges to isotropic Gaussian noise.

To turn this noising process into a generative model, diffusion models learn an approximate \emph{reverse} process that reconstructs samples from noise back to data.
A neural network $\varepsilon_\theta(x_t, t)$ is trained to predict the injected noise $\varepsilon$ given a noisy sample $x_t$ and a timestep embedding $t$.
The standard training objective is
\begin{equation}
  \mathcal{L}
  = \mathbb{E}_{x_0,\, t,\, \varepsilon}
    \Big[
      \big\|
        \varepsilon_\theta(x_t, t) - \varepsilon
      \big\|_2^2
    \Big],
\end{equation}
which encourages $\varepsilon_\theta$ to recover the noise that produced $x_t$ from $x_0$.
At inference time, one initializes $x_T$ from a standard Gaussian and iteratively applies the reverse process parameterized by $\varepsilon_\theta$ to obtain a synthetic sample $\hat{x}_0$ from the learned distribution.

\section{\pjn Details} \label{app:method_details}

\textbf{About Item Matching.}
In \S\ref{sec:item_matching}, we describe the item matching procedure which is designed to construct slot-based, item-aligned bin states.
Specifically, we formulate item matching as a Hungarian matching problem with priors derived from the stow behavior, in which the cost function is defined as,
\begin{equation}
  C_{ij}
  = (1-D_{ij})
  + \lambda_{\text{pos}} \, \Phi_{\text{pos}}(i,j)
  + \lambda_{\text{ord}} \, \Phi_{\text{ord}}(i,j),
\end{equation}
where $D_{ij}$ is pairwise cosine similarity between instance visual embeddings, and $\Phi_{\text{pos}}$ and $\Phi_{\text{ord}}$ encode the placement and order-preservation penalties, respectively. We solve the linear assignment problem defined by $C_{ij}$ using the Hungarian algorithm to obtain the final one-to-one correspondence between pre- and post-stow instances.
For completeness, we describe the concrete form of the placement and order-preservation penalties used in the matching cost.

\textit{Position Prior.}
Let $\{m_i^{\text{pre}}\}_{i=1}^{N_{\text{pre}}}$ and $\{m_j^{\text{post}}\}_{j=1}^{N_{\text{post}}}$ denote the instance masks in the pre- and post-stow snapshots, respectively,
and let $c_j^{\text{post}} \in \mathbb{R}^2$ be the center of the $j$-th post-stow instance (e.g., the center of its bounding box), and $p_{\text{ins}} \in \mathbb{R}^2$ and $p_{\text{push}} \in \mathbb{R}^2$ denote that of the planned insertion and sweeping positions from the stow intent.
We first compute a distance to the intended placement region,
\begin{equation}
  d_j^{\text{pos}}
  = \min\bigl(
      \|c_j^{\text{post}} - p_{\text{ins}}\|_2,\,
      \|c_j^{\text{post}} - p_{\text{push}}\|_2
    \bigr),
\end{equation}
where the term involving $p_{\text{push}}$ is omitted if no sweeping is performed.
We then convert this distance into a penalty
\begin{equation}
  \Phi_{\text{pos}}(i,j)
  = \exp\!\left(-\frac{d_j^{\text{pos}}}{\sigma_{\text{pos}}}\right),
\end{equation}
where $\sigma_{\text{pos}}$ controls the spatial decay.
This term penalizes matching pre-existing items to post-stow instances that lie very close to the intended insertion region, which are more likely to be the newly stowed item.

\textit{Order-Preservation Prior.}
Let $x_i^{\text{pre}}$ and $x_j^{\text{post}}$ denote the normalized horizontal coordinates of the pre- and post-stow instances (e.g., the centers normalized to $[0,1]$ along the bin width).
We obtain left-to-right ranks $r_i^{\text{pre}}$ and $r_j^{\text{post}}$ by sorting instances by $x$, and normalize them to $[0,1]$,
\begin{equation}
  \hat r_i^{\text{pre}} = \frac{r_i^{\text{pre}} - 1}{N_{\text{pre}} - 1},
  \qquad
  \hat r_j^{\text{post}} = \frac{r_j^{\text{post}} - 1}{N_{\text{post}} - 1}.
\end{equation}
The order-preservation penalty for a candidate match $(i,j)$ is then
\begin{equation}
  \Phi_{\text{ord}}(i,j)
  = \bigl|\hat r_i^{\text{pre}} - \hat r_j^{\text{post}}\bigr|,
\end{equation}
which discourages assignments that significantly change the left--right order of pre-existing items. 
When only one item is in the pre-stow state, the order-preservation prior will be skipped.

\textbf{More Clarifications.}
We clarify two practical aspects of the data processing steps introduced in \S\ref{sec:item_matching}.

First, both the instance mask extraction and the item matching procedure are imperfect and can introduce noise into the supervision.
Quantifying this noise would require ground-truth instance masks and correspondences, which is infeasible to obtain at scale.
Instead, we apply simple post-processing to suppress obvious errors.
For each stow event, the number of post-stow instances must equal the number of pre-stow instances plus one. If this constraint is violated, we discard the stow as a likely failure of the instance segmentation step.
For the remaining stows, we further audit the one-to-one matching with a large vision-language model, Claude-3.7~\cite{claude} in a visual question answering setup, which inspects the pre- and post-stow RGB images and judges whether the proposed correspondences are consistent.
Some residual noise may remain, and we treat it as label noise in the training data.

Second, the canonical new-item mask extracted from the ground-truth post-stow masks is used in both training and evaluation.
This choice reflects our focus on evaluating how well the model can predict post-stow layouts when given the in-bin contact-surface view of the new item.
In \dataset, only an induct-view image of the new item is available, which does not directly correspond to the surface that faces outward once the item is placed in the bin. Using the canonical mask derived from ground-truth post-stow masks therefore serves as a proof-of-concept proxy for this view.
When targeting for a real deployed system, one could obtain the required view by adding a camera that observes the item surface that will face outward in the bin, or by learning a separate model that predicts this view from the induct observation, for example via 3D reconstruction~\cite{yang2023sam3d}.
Designing such a module is beyond the scope of this work and we leave it to future research.

\section{More Experimental Settings} \label{app:exp_settings}

\begin{wraptable}{r}{0.3\textwidth}
\vspace{-10pt}
\centering
\caption{Dataset statistics} \label{tab:data}
\small
\tabcolsep=0.1cm
\vspace{-5pt}
\begin{tabular}{c|cc|cc}
\toprule
Intent                 & \multicolumn{2}{c|}{Direct Insert} & \multicolumn{2}{c}{Sweep Insert} \\ \midrule
\multirow{2}{*}{Split} & Train           & Test            & Train          & Test         \\
                       & 26,531          & 6,558          & 20,678         & 5,350        \\ \bottomrule
\end{tabular}
\vspace{-10pt}
\end{wraptable}
\textbf{About Dataset Statistics.}
Table~\ref{tab:data} summarizes the data statistics. As described in \S\ref{sec:exp_setup}, we focus on successful stow events in \dataset and group them by stow intent into two modes. \emph{Direct insert} only execute an insertion action, and \emph{sweep insert} uses an additional sweeping action to create space. Within each mode we randomly split stows into training and test sets using an 8:2 ratio.

\textbf{About the Downstream Evaluations.}
Here we provide additional details for the two downstream evaluations, Delta Linear Opportunity (DLO) prediction and multi-stow reasoning.

For DLO prediction, we train a vision-based DLO predictor that takes as input the pre-stow bin RGB image, the pre-stow instance masks, and a candidate post-stow instance mask.  
Each of the three inputs is encoded separately by a frozen DINOv2 backbone~\cite{oquab2023dinov2}, yielding three sets of tokens.
We then add a learnable input-type embedding to the tokens from each stream so that the model can distinguish them, and concatenate all tokens along the sequence dimension.
A learnable class token is further concatenated to this sequence and passes through two transformer blocks with self-attention, which fuse information across all inputs.  
Finally, the class token output is fed to a small MLP head that produces a single scalar, the predicted DLO.
This predictor is trained with ground-truth post-stow masks so that it learns the best achievable mapping from the combination of pre-stow state and true post-stow state to DLO, and the post-stow mask is only swapped at test time when comparing different synthetic post-stow state.

For multi-stow reasoning, \pjn requires a pre-stow RGB image at each step, but only the first pre-stow RGB is directly available. To obtain RGB inputs for later steps, we synthesize post-stow RGB images from the predicted post-stow masks.  
Specifically, we first extract the item texture from its induct RGB view provided in \dataset by segmenting out the new item and removing its background.  
Given a predicted post-stow mask for the new item in the bin, we then paste this texture into the pre-stow RGB image at the mask location, producing a synthetic post-stow RGB image.   This synthesized image serves as the pre-stow RGB input for the next stow in the chain during multi-stow rollout.

\textbf{About Fine-Grained IoU Analysis.}
For the fine-grained IoU analysis in \S\ref{sec:exp_direct}, we further partition test stows along two axes, the size of the newly inserted item and the difficulty of the stow derived from the copy-paste baseline.

\begin{figure}[htb]
\vspace{-10pt}
    \centering
    \begin{subfigure}[b]{0.48\linewidth}
        \centering
        \includegraphics[width=.48\linewidth]{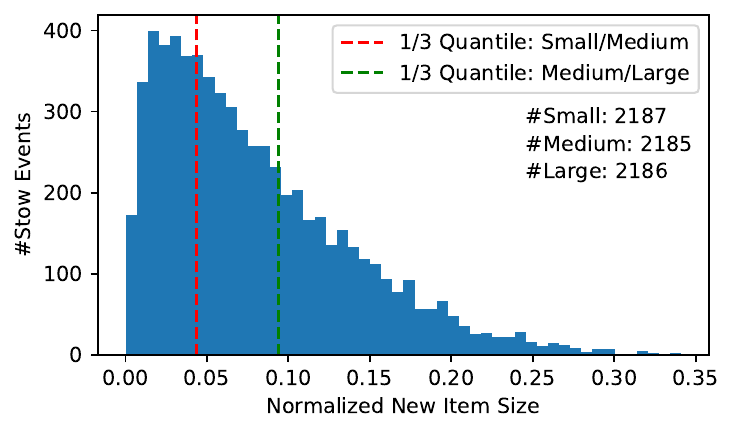}
        \includegraphics[width=.48\linewidth]{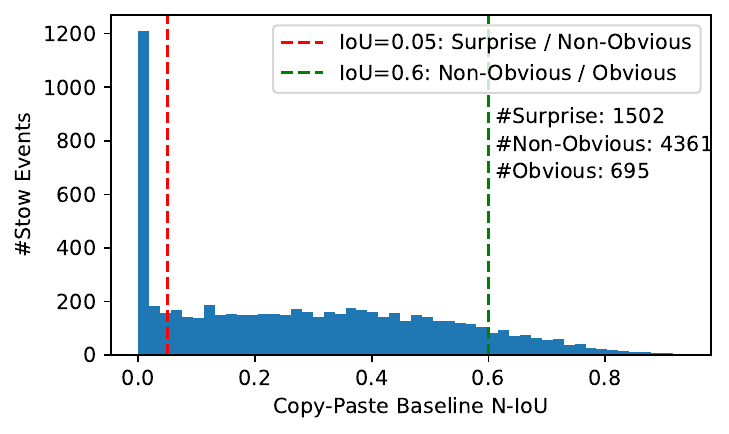} 
        \caption{Direct Insert}
    \end{subfigure}%
    \hfill
    \begin{subfigure}[b]{0.48\linewidth}
        \centering
        \includegraphics[width=.48\linewidth]{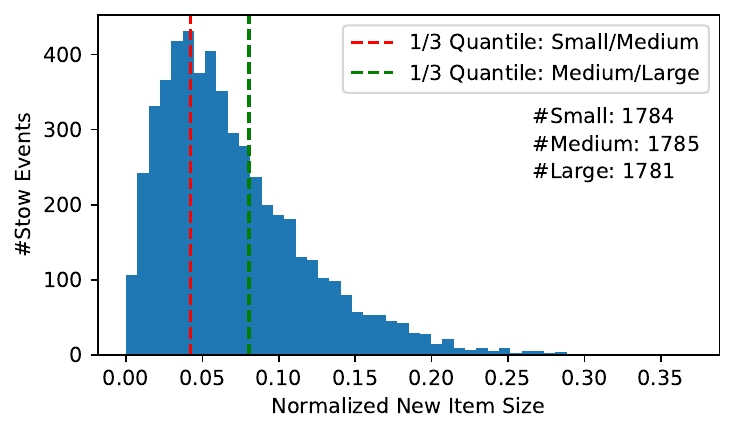}
        \includegraphics[width=.48\linewidth]{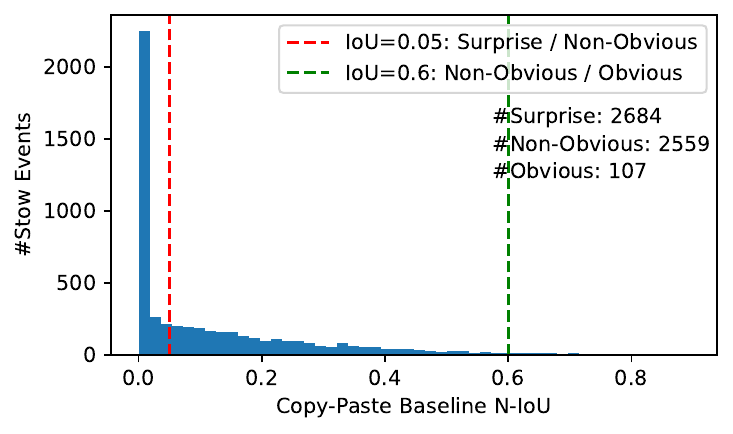} 
        \caption{Sweep Insert}
    \end{subfigure}
    \caption{Fined-grained IoU partition according to the normalized new item size (left) and the difficulty of the stow events derived from the performance of the copy-paste baseline (right) for (a) direct insert and (b) sweep insert modes respectively.}
    \label{fig:IoU-partition}
    \vspace{-5pt}
\end{figure}

For the size buckets, we measure the normalized new-item size as the area of the new-item mask divided by the bin area in the stow view. Within each stow mode , we compute the $1/3$ and $2/3$ quantiles of this normalized size distribution.  
Stows with size below the first quantile are labeled small, those between the first and second quantiles are medium, and those above the second quantile are large.  
This evenly spaced quantile partition yields three groups of comparable size and isolates very small items, for which IoU is particularly sensitive to small spatial errors. 
The left plots in Figure~\ref{fig:IoU-partition} visualize this partition, along with the number of stows in each bucket.

For the difficulty buckets, we instead threshold the N-IoU achieved by the copy-paste baseline.  
We run this naive baseline on all test stows and collect its N-IoU scores. Within each stow mode, stows are then grouped by these scores into three difficulty levels.
Specifically, stows with baseline N-IoU below $0.05$ are surprising cases, where the naive copy-paste rarely places the new item correctly and often yields near-zero overlap.  
Stows with N-IoU between $0.05$ and $0.60$ are non-obvious cases, where the baseline achieves partial overlap but still leaves room for improvement.  
Finally, stows with N-IoU above $0.60$ are labeled obvious cases, where the simple strategy already performs well.
The right plots in Figure~\ref{fig:IoU-partition} show the N-IoU histograms and the counts in each difficulty bucket.

\section{More Visualizations} \label{app:vis}
\begin{figure}[htb]
    \centering
    \includegraphics[width=.93\textwidth]{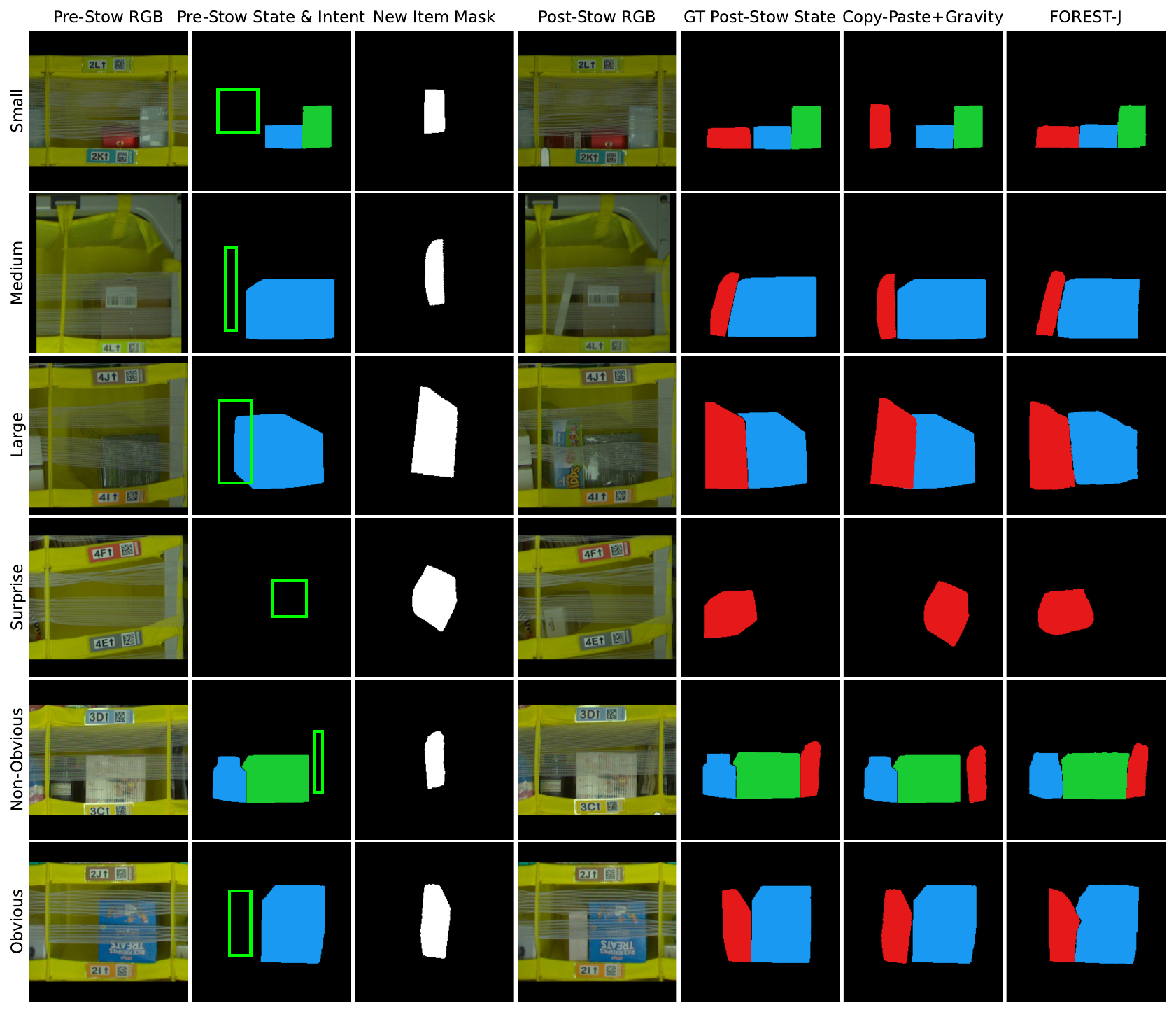}
    \vspace{-.2cm}
    \caption{More examples of post-stow bin state prediction for \textbf{direct insert} stow events. Representatives are selected from stows with small, medium, large item or surprise, non-obvious, obvious stow events.}
    \label{fig:direct_vis_app1}
\end{figure}

\begin{figure}[htb]
    \centering
    \includegraphics[width=.93\textwidth]{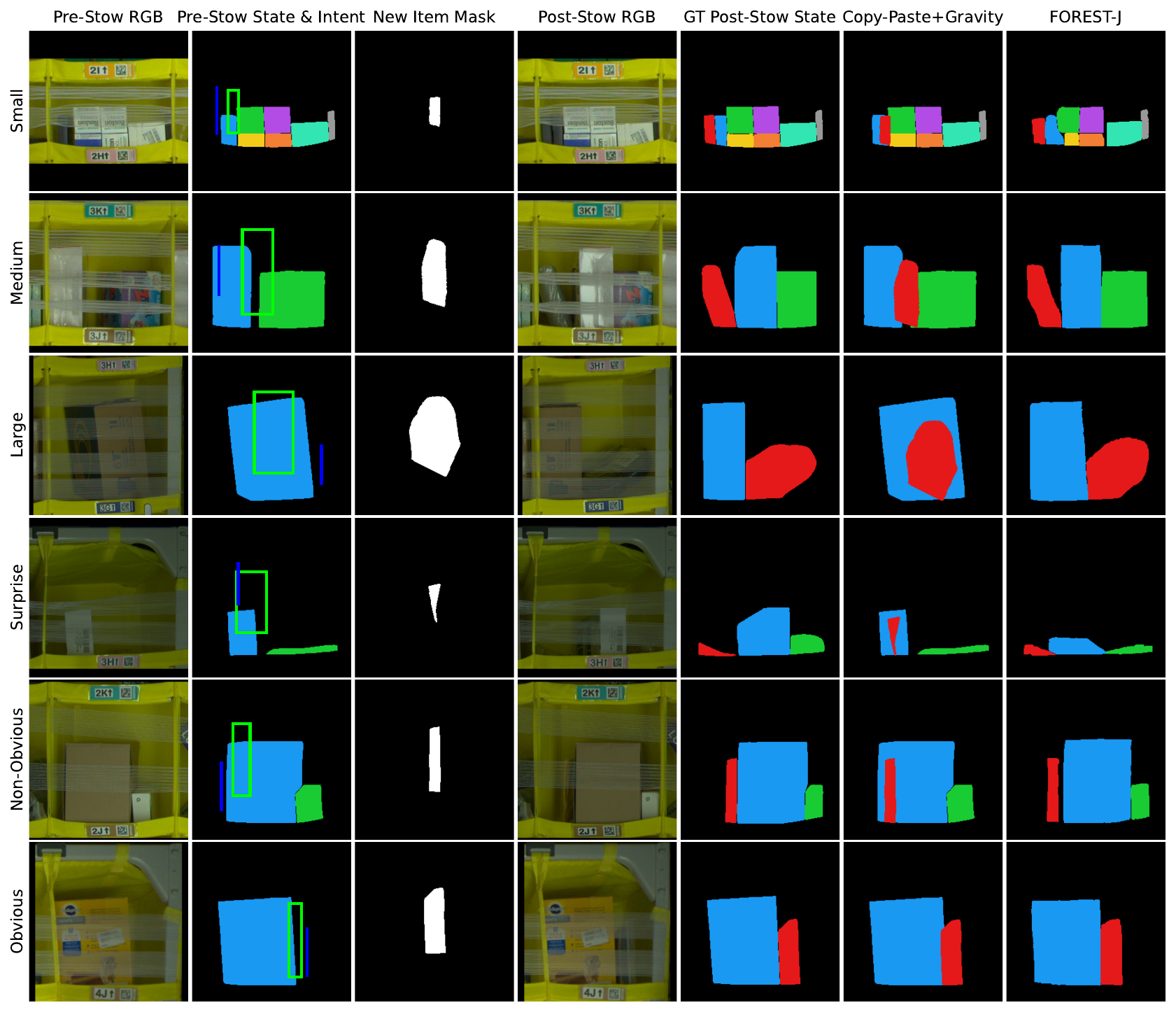}
    \vspace{-.2cm}
    \caption{More examples of post-stow bin state prediction for \textbf{sweep insert} stow events. Representatives are selected from stows with small, medium, large item or surprise, non-obvious, obvious stow events.}
    \label{fig:direct_vis_app2}
\end{figure}

\textbf{Single-Step Stow.}
Figure~\ref{fig:direct_vis_app1} and Figure~\ref{fig:direct_vis_app2} provide additional single-step qualitative examples for direct insert and sweep insert stows.  
Each row shows one stow event, with columns corresponding to pre-stow RGB, pre-stow bin state with stow intent, new item mask, post-stow RGB, ground-truth post-stow bin state, copy-paste with gravity, and \pjn-J predictions.
In the two visualizations, we select representative stows from the small, medium, and large new-item size buckets and from the surprise, non-obvious, and obvious difficulty groups.  
\pjn-J consistently produces post-stow layouts that closely align with the ground-truth masks, while copy-paste with gravity often misses item tipping, sliding, or coordinated motion of pre-existing items.

\textbf{Multi-Step Stow.}
Table~\ref{tab:multistow} reports per-step N-IoU of predicted post-stow masks along multi-stow chain.
For the copy-paste-with-gravity (CP+G) baseline, N-IoU drops quickly below $0.2$ from step~3 onward, showing that this heuristic cannot support long-horizon rollouts.
In contrast, both \pjn-DI+\pjn-SI and \pjn-J maintain much higher N-IoU under rollout, and their teacher-forcing curves stay relatively stable across steps.
The gap between teacher-forcing and rollout therefore reflects accumulated prediction error.

\begin{table}[htb]
\centering
\caption{N-IoU of predicted post-stow masks along multi-stow chains (table version of Figure~\ref{fig:multistow}). We compare \pjn-J and the intent-specific models, \pjn-DI+\pjn-SI, under teacher-forcing and rollout modes, together with the copy-paste-with-gravity (CP+G) baseline. The number of stows for each step are included as reference.} \label{tab:multistow}
\scriptsize
\tabcolsep=0.1cm
\vspace{-3pt}
\begin{tabular}{cc|ccccc}
\toprule
\multicolumn{2}{c|}{Multi-Step Chain} & \multicolumn{5}{c}{N-IoU}                                                                                                            \\
Step             & Count             & CP + G & \pjn-DI + \pjn-SI (Rollout) & \pjn-DI + \pjn-SI (Teacher-Forcing) & \pjn-J (Rollout) & \pjn-J (Teacher-Forcing) \\ \midrule
1                & 1581              & 0.2868 & 0.6934                          & 0.6911                                  & 0.6992             & 0.6993                     \\
2                & 1581              & 0.2353 & 0.6069                          & 0.6478                                  & 0.6211             & 0.6606                     \\
3                & 189               & 0.2    & 0.4755                          & 0.5590                                  & 0.4980             & 0.5810                     \\
4                & 16                & 0.1393 & 0.4458                          & 0.5773                                  & 0.3871             & 0.5471                     \\
\bottomrule
\end{tabular}
\end{table}

\begin{figure}[t]
    \centering
    \includegraphics[width=0.6\textwidth]{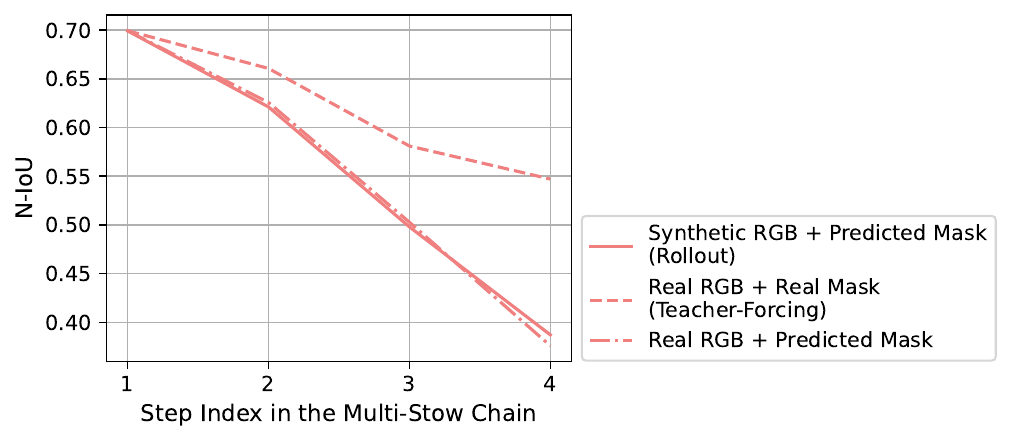}
    \caption{
    We compare N-IoU along multi-stow chains under three settings for \pjn-J:
    rollout with synthetic RGB and predicted masks, teacher forcing with real RGB and ground-truth masks, and rollout-style evaluation with real RGB and predicted masks.
    }
    \label{fig:multi_stow_rgb_ablation}
\end{figure}

One concern in the rollout evaluation is that the synthesized pre-stow RGBs may also introduce error, which could confound the analysis of error accumulation from the predicted post-stow masks.
To isolate this effect, we run an additional multi-stow experiment where we keep the pre-stow RGB images \emph{real} at every step, but feed the \emph{predicted} post-stow masks from \pjn-J as the bin-state input for the next step.
Figure~\ref{fig:multi_stow_rgb_ablation} compares three settings, (i) rollout with synthetic RGB and predicted masks, (ii) teacher forcing with real RGB and ground-truth masks, and (iii) rollout-style evaluation with real RGB and predicted masks.
The curves for (i) and (iii) are almost identical across steps, while both remain clearly below (ii), indicating that the degradation comes primarily from accumulating errors in the predicted masks rather than from the synthesized RGB.

\begin{figure}[htb]
  \centering
  \includegraphics[width=\linewidth]{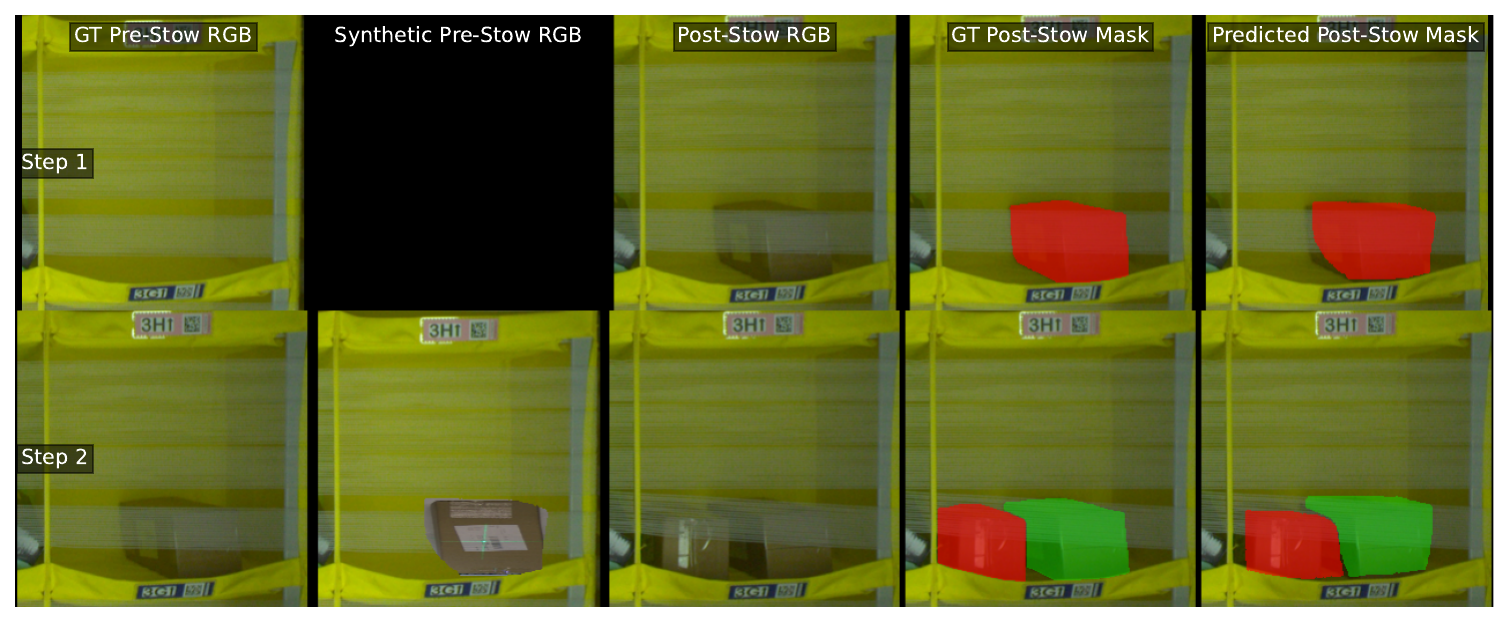}
  \caption{
    Multi-step stow examples (Ex.1).
    Each row is one stow in a chain and each column shows, from left to right:
    ground-truth pre-stow RGB, synthesized pre-stow RGB, ground-truth post-stow RGB,
    ground-truth post-stow mask, and \pjn-J predicted post-stow mask. 
    In the post-stow mask, the red mask represents the new item, while the green one are pre-existing items.
  }
  \label{fig:multistep_vis1}
\end{figure}

\begin{figure}[htb]
  \centering
  \includegraphics[width=\linewidth]{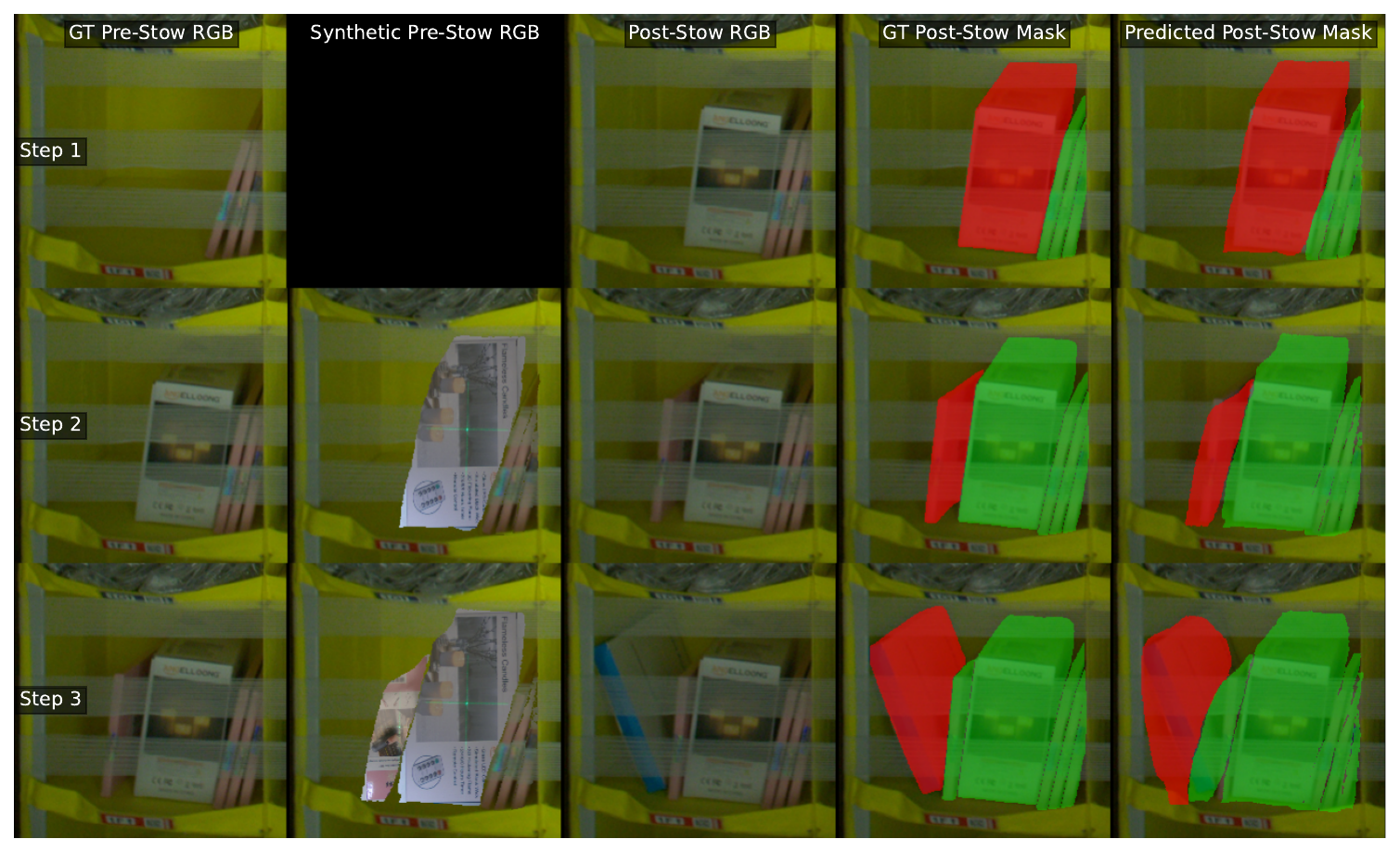}
  \caption{More multi-step stow examples (Ex.2).}
  \label{fig:multistep_vis2}
\end{figure}

\begin{figure}[htb]
  \centering
  \includegraphics[width=\linewidth]{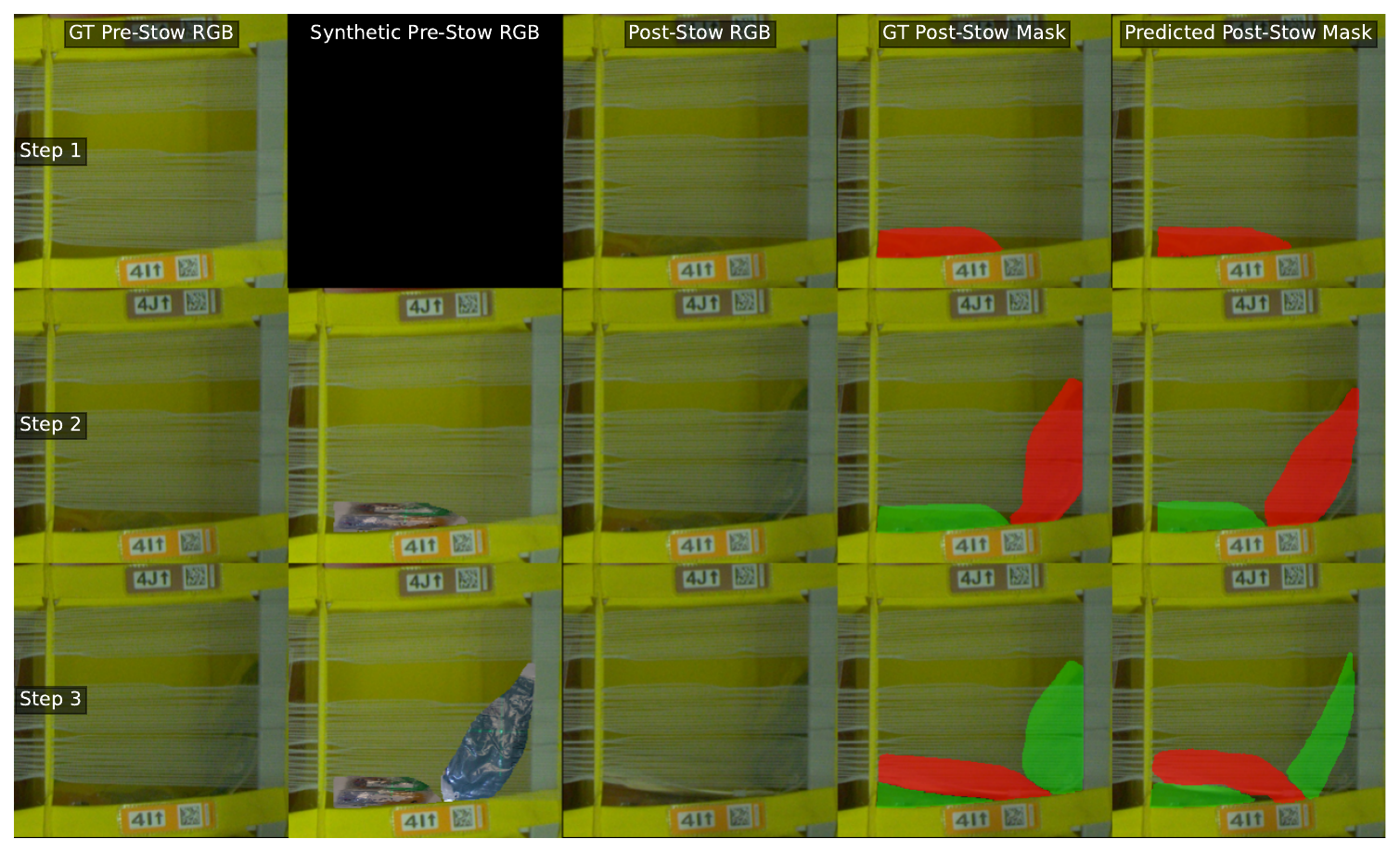}
  \caption{More multi-step stow examples (Ex.3).}
  \label{fig:multistep_vis3}
\end{figure}

\begin{figure}[htb]
  \centering
  \includegraphics[width=\linewidth]{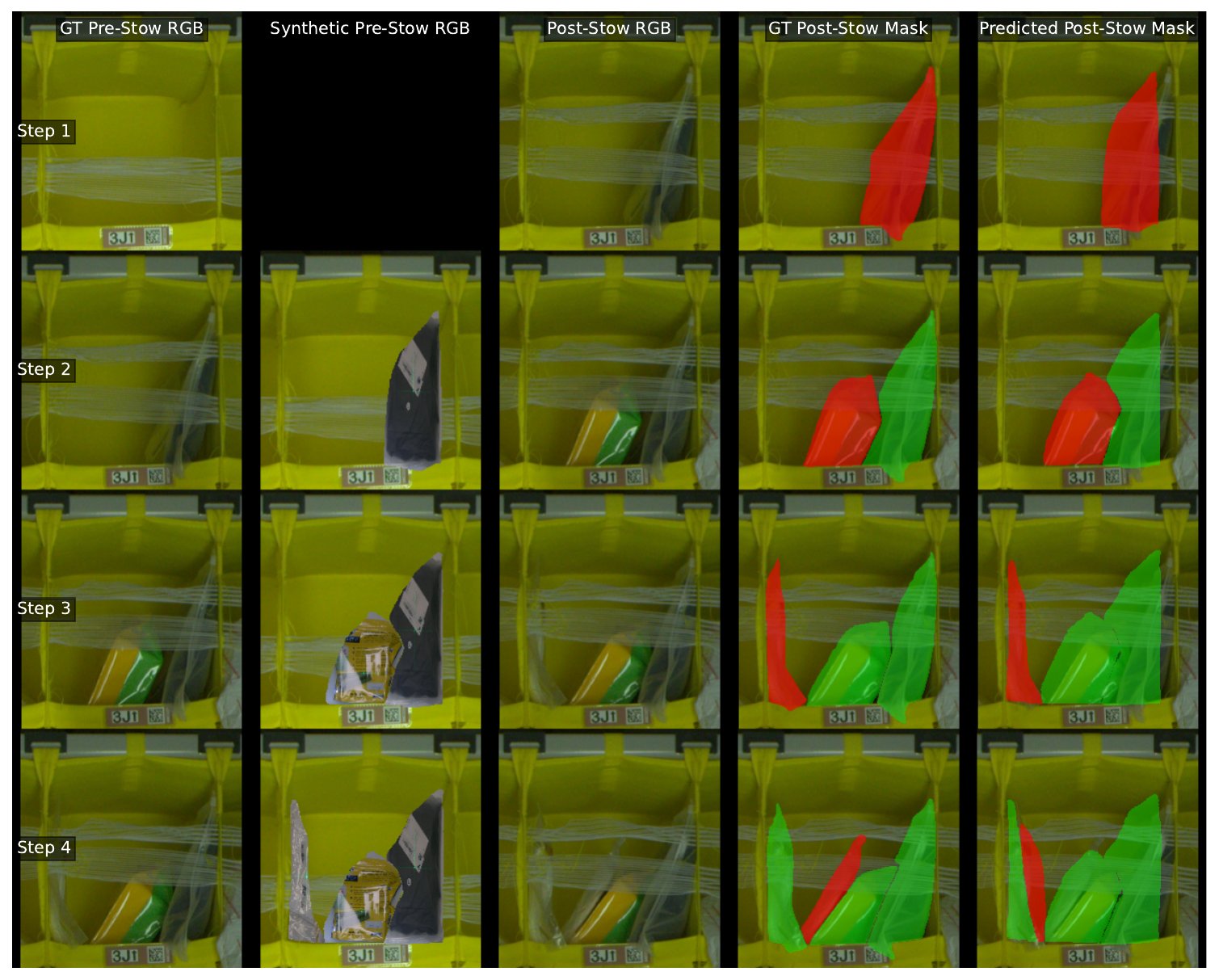}
  \caption{More multi-step stow examples (Ex.4; long chain case with last step not accurate).}
  \label{fig:multistep_vis4}
\end{figure}

\begin{figure}[htb]
  \centering
  \includegraphics[width=\linewidth]{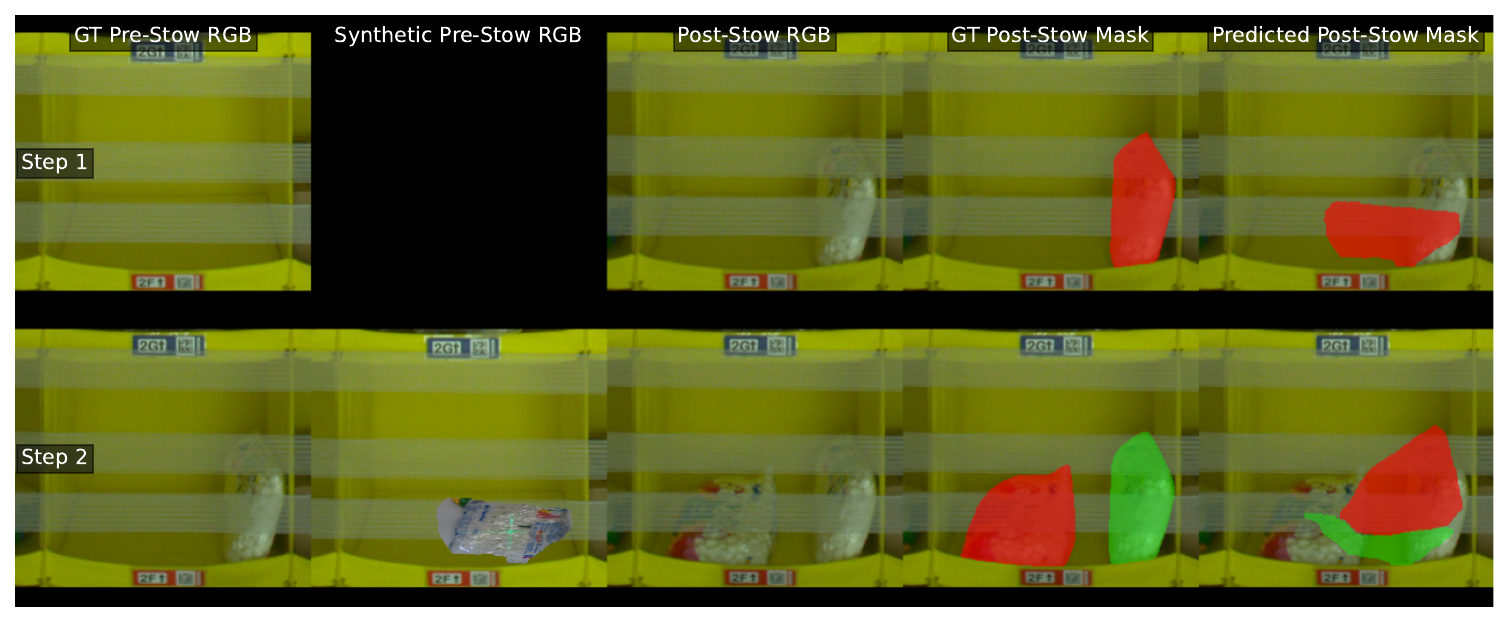}
  \caption{More multi-step stow examples (Ex.5; failure case).}
  \label{fig:multistep_vis5}
\end{figure}

\begin{figure}[htb]
  \centering
  \includegraphics[width=\linewidth]{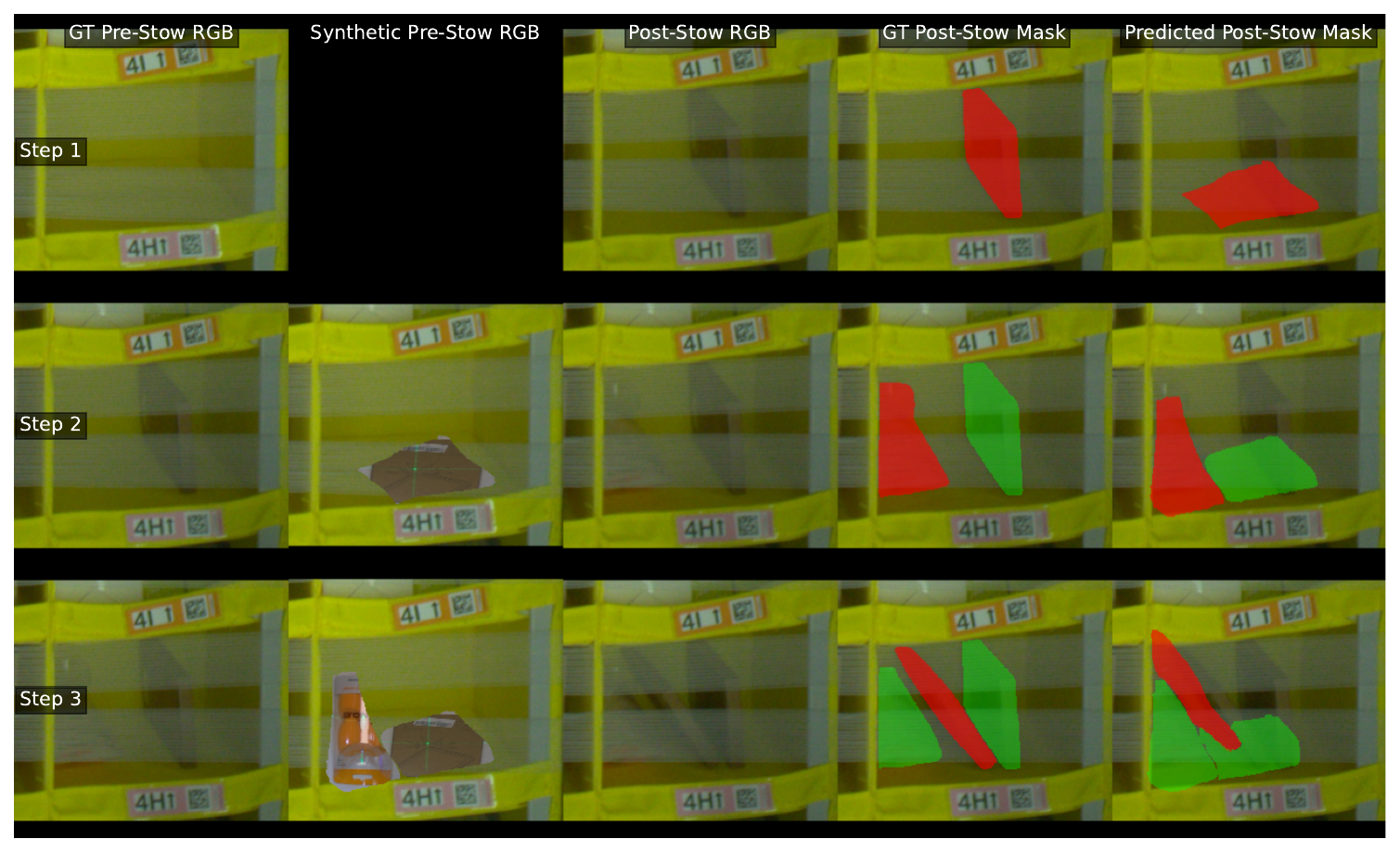}
  \caption{More multi-step stow examples (Ex.6; failure case).}
  \label{fig:multistep_vis6}
\end{figure}

Figures~\ref{fig:multistep_vis1} to \ref{fig:multistep_vis6} provide additional visualizations.
Each figure shows one multi-stow chain, with one row per stow step.
From left to right, we display the ground-truth pre-stow RGB image, the synthesized pre-stow RGB image used as input to the next step, the ground-truth post-stow RGB image, the ground-truth post-stow mask, and the \pjn-J predicted post-stow mask.
These examples include both typical successes, where the model tracks how items accumulate and reconfigure over several stows, and failure cases, where unstable items lead to larger drift in the rolled-out predictions (Figures~\ref{fig:multistep_vis5} and \ref{fig:multistep_vis6}).

\clearpage
\section{Ablation Study on Conditions} \label{sec:exp_ablations}
We ablate the conditioning signals used by \pjn, namely the pre-stow RGB image and the item property token, on direct insert stow events.
Table~\ref{tab:abl_cond} reports N-IoU and O-IoU, together with fine-grained N-IoU by item size and stow difficulty.
Overall, removing either signal degrades performance, and removing both yields the largest drop. N-IoU falls from $0.701$ to $0.579$ without RGB, to $0.661$ without the property token, and to $0.509$ when both are removed.
The degradation is especially pronounced for small and difficult stows, suggesting that these challenging cases benefit most from richer conditioning.

\begin{table}[htb]
\centering
\caption{Ablation on conditioning signals for \pjn-DI on \emph{direct insert} stows.  We report N-IoU for the newly inserted item and O-IoU for pre-existing items, 
together with fine-grained N-IoU by new-item size and stow difficulty.} \label{tab:abl_cond}
\small
\tabcolsep=0.08cm
\vspace{-5pt}
\begin{tabular}{lcccc}
\toprule
& \textbf{\pjn-DI} & \textbf{w/o RGB} & \textbf{w/o Prop} & \textbf{w/o RGB\,+\,Prop} \\
\midrule
\multicolumn{5}{c}{\emph{Overall}} \\ 
N-IoU & \textbf{0.7017} & 0.5799 & 0.6619 & 0.5089 \\
O-IoU & \textbf{0.8550} & 0.8232 & 0.8321 & 0.7675 \\
\midrule
\multicolumn{5}{c}{\emph{N-IoU by item size}} \\
Small  & \textbf{0.5771} & 0.4549 & 0.5215 & 0.3799 \\
Medium & \textbf{0.7263} & 0.6093 & 0.6881 & 0.5311 \\
Large  & \textbf{0.8021} & 0.6756 & 0.7764 & 0.6158 \\
\midrule
\multicolumn{5}{c}{\emph{N-IoU by difficulty}} \\
Surprises    & \textbf{0.5623} & 0.4238 & 0.5042 & 0.3448 \\
Non-obvious  & \textbf{0.7354} & 0.6186 & 0.7004 & 0.5468 \\
Obvious      & \textbf{0.7920} & 0.6745 & 0.7619 & 0.6257 \\
\bottomrule
\end{tabular}
\end{table}

\section{Limitations}~\label{app:limitation}
This work has several limitations. 
\pjn focuses on single-step transitions and mask-based bin states, leaving richer visual channels and multi-step training for future exploration. 
Moreover, due to limitations of \dataset, our new-item representation relies on a canonical mask extracted from the ground-truth post-stow view rather than a true contact-surface view. 
Reproducing \pjn using contact surfaces derived from induct views is an important next step to better match deployment conditions. 
Addressing these limitations and coupling \pjn more tightly with stow policy learning and planning are promising directions for future work.


\end{document}